\PassOptionsToPackage{table}{xcolor}
\PassOptionsToPackage{framemethod=tikz}{mdframed}
\documentclass[11pt,a4paper]{deepseek}
\usepackage[authoryear,sort&compress,round]{natbib}
\usepackage[utf8]{inputenc}
\usepackage[T1]{fontenc}
\usepackage{amsmath,amssymb,booktabs,tabularx,array}
\usepackage{longtable}
\usepackage{graphicx,xcolor,microtype}
\usepackage{subcaption}
\usepackage{capt-of}
\usepackage{wrapfig}
\usepackage{needspace}
\usepackage{colortbl}
\definecolor{paperAccent}{HTML}{406D5C}
\definecolor{paperAccentSoft}{HTML}{EEF3F0}

\usepackage{algorithm,algpseudocode}
\usepackage{tikz}
\usetikzlibrary{arrows.meta,positioning,fit,backgrounds}
\usepackage{xurl}
\usepackage{xspace}
\usepackage{hyperref}

\hypersetup{colorlinks=true,linkcolor=blue!45!black,citecolor=blue!45!black,urlcolor=blue!45!black}
\newcommand{\method}{{\fontfamily{ppl}\selectfont uni-robot-agent}\xspace}

\ifXeTeX
  \RequirePackage{fontspec}
\fi
\fancypagestyle{firststyle}{%
  \fancyhf{}%
  \fancyhead[L]{\includegraphics[height=22pt]{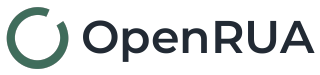}}%
}
\RequirePackage{etoolbox}
\RequirePackage{fontawesome}
\patchcmd{\maketitle}{\vspace*{3pt}}{\vspace*{-16pt}}{}{%
  \PackageError{deepseek-settings}{Title offset patch failed}{Check deepseek.cls.}}
\patchcmd{\maketitle}{\vskip20pt}{\vskip4pt}{}{%
  \PackageError{deepseek-settings}{Title spacing patch failed}{Check deepseek.cls.}}
\patchcmd{\maketitle}{\vskip30pt}{\vskip0pt}{}{%
  \PackageError{deepseek-settings}{Author spacing patch failed}{Check deepseek.cls.}}
\renewcommand\Affilfont{\normalfont\fontsize{11}{15}\selectfont\centering}
\makeatletter
\renewcommand{\abscontent}{%
  \par{\centering\normalfont\fontsize{11}{13}\selectfont
    \raisebox{-1.5pt}{\includegraphics[height=10pt]{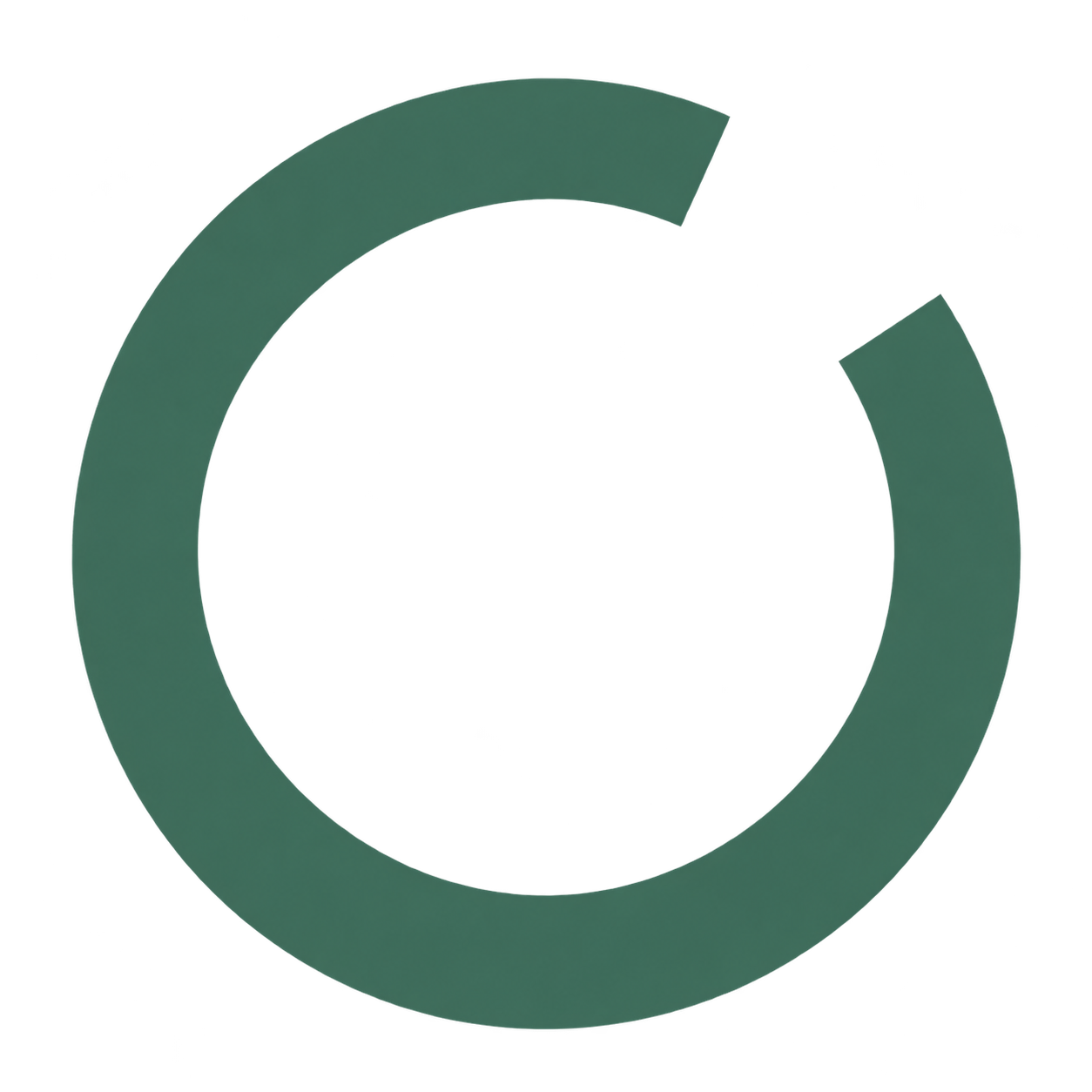}}\hspace{0.35em}\href{https://openrua.ai/uni-robot-agent}{Blog}\quad\textbar\quad
    \raisebox{-1.5pt}{\resizebox{!}{10pt}{\raisebox{\depth}{\textcolor{black}{\faGithub}}}}\hspace{0.35em}\href{https://github.com/Open-RUA/uni-robot-agent}{Code}\quad\textbar\quad
    \raisebox{-1.5pt}{\resizebox{!}{10pt}{\raisebox{\depth}{\textcolor{black}{\faGlobe}}}}\hspace{0.35em}\href{https://uni-robot-agent.com}{Website}\quad\textbar\quad
    \raisebox{-1.5pt}{\resizebox{!}{10pt}{\raisebox{\depth}{\textcolor{black}{\faBook}}}}\hspace{0.35em}\href{https://uni-robot-agent.com/docs}{Documentation}\par}%
  \vspace{0.5ex}%
  \par{\centering\fontsize{15pt}{14pt}\selectfont\textbf{Abstract}\par}%
  \vspace{1ex}%
  \noindent\parbox{\linewidth}{\absfont\theabstract}\par
  \@ifundefined{@keywords}{}{\vskip1em\noindent\keywordsfont Keywords: \@keywords}%
}
\makeatother
\ifXeTeX
  \microtypesetup{tracking=false}
\fi

\title{\centering Democratizing Robot-Use Agents:\\A Unified Harness with Open Models}
\appto\Authfont{\bfseries}
\author{%
Jiarui Yuan$^*$,
Yuxin Zuo$^{*\dagger}$,
Xiaoyu Huang$^{*\dagger}$,
Wenbo Zhang$^*$,
Kwanyoung Park$^*$,\authorcr
Junlin Yang$^*$,
Tianxiang Zhang$^*$,
Ruoqu Chen$^*$,
Yutao Ouyang$^*$,
Bingxiang He$^*$,
Xingtai Lv$^*$,\authorcr
Hongjin Wu$^*$,
Jiajun Liu,
Jiahao Zhao,
Tianwei Luo,
Haozhan Li,
Cheng Qian,
Jiaxin Ge,\authorcr
Huan-ang Gao,
Kaisen Yang,
Jiaqi Deng,
Qixin Xu,
Siqiao Huang,
Bowen Zhou,\authorcr
Mengdi Xu,
Zhiyuan Liu,
Pieter Abbeel,
Ning Ding\\
\normalfont Tsinghua University\qquad UC, Berkeley\\
\vspace{6pt}
\normalfont\small $^*$Core Contributor.\qquad$^\dagger$Project Lead.
}
\date{2026-11-02}

\begin{abstract}
Robot-use agents (RUAs) have shown promising zero-shot manipulation capabilities, but reliance on frontier, proprietary models limits research into further training.
To democratize RUA research, we introduce \method, a unified harness for competitive open-weight robot use that supports further training. Without privileged inputs, an out-of-the-box Qwen3.8-Flash-Next achieves a RoboDojo Score of 31.39, surpassing the published GPT-6 Astra–RoboProbe result of 28.97 and many VLA/WAM baselines, and performs competitively on 19 real-world tasks.
Our analysis across model generations, families, and sizes links effective robot use to four general capabilities, suggesting that future RUA training should develop them jointly.
\end{abstract}

\begin{document}
% Give the title page a smaller top margin; restore the class layout for the body.
\newgeometry{left=25mm,right=25mm,top=22mm,bottom=30mm,headheight=40pt,headsep=20pt}
\maketitle

\begin{figure}[H]
\centering
\includegraphics[width=.9\linewidth]{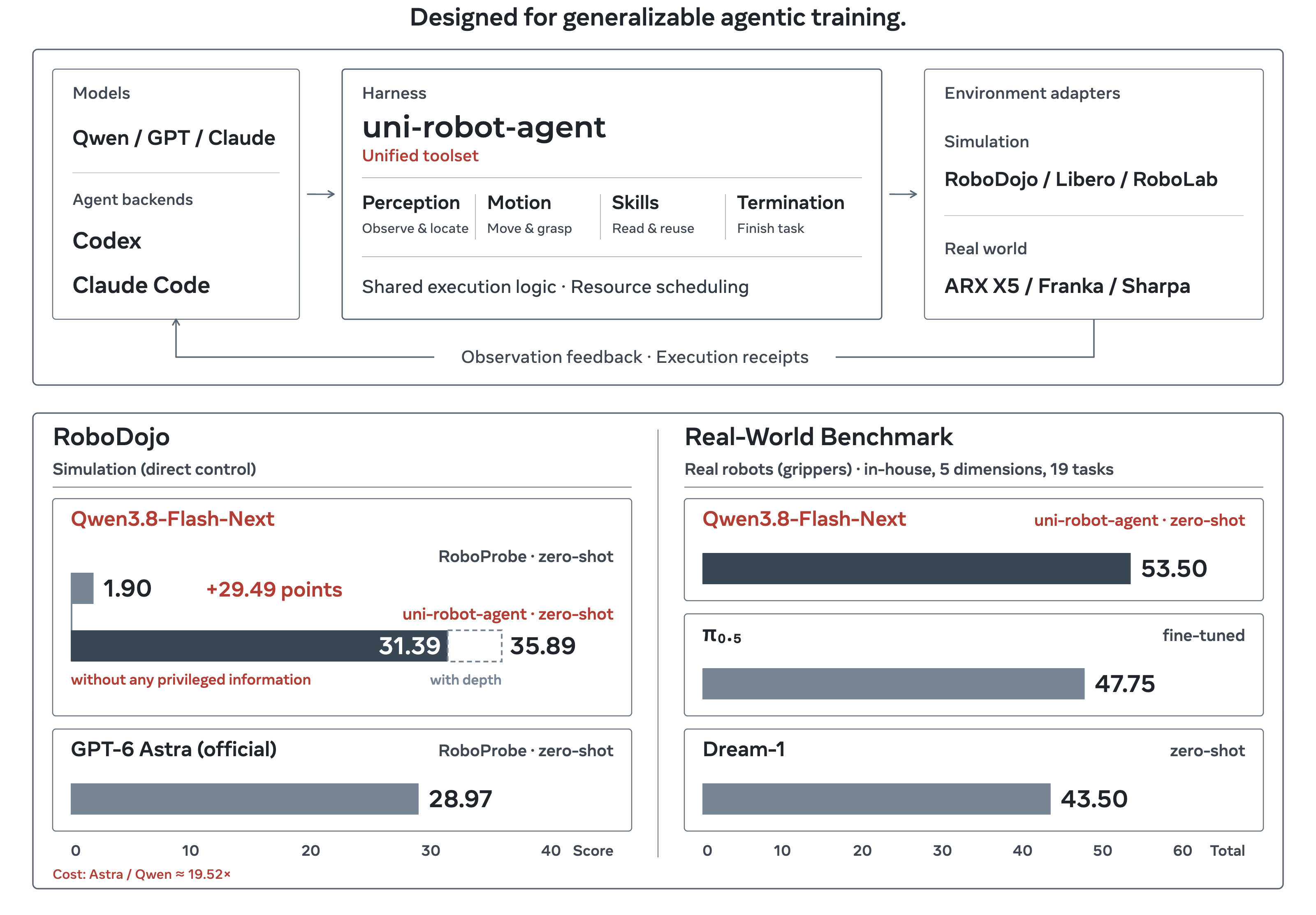}
\caption{\textbf{\method: results and system overview.}
Top: shared tools, execution, and feedback connect models to environment adapters, supporting generalizable agentic training across environments.
Bottom: Simulation and real-world scores.
% The $\pi_{0.5}$ baseline is fine-tuned with 10 hours of robot data per task.
}
\label{fig:overview}
\end{figure}

\restoregeometry
\begingroup
\setcounter{tocdepth}{2}
\tableofcontents
\endgroup
\clearpage
\section{Introduction}
\label{sec:introduction}

Recent breakthroughs in frontier foundation models (e.g., GPT-6 Astra~\citep{openai2026astra}, Opus 5.5~\citep{opus}) have shown surprising zero-shot performance in robotic manipulation, highlighting robot-use agents (RUAs)~\citep{isola2026robotuse} as a promising approach for robotics~\citep{rua1,rua2,rua3}.
This approach effectively transfers the models' generic problem-solving capabilities, such as task decomposition and failure recovery, which are hard to learn from limited robotics data alone, yet extremely useful for robotics~\citep{isola2026robotuse,chu2026openrua,jia2026agent}.

%For example, GPT-6 Astra outperformed 40 public policies in an early zero-shot evaluation covering all 42 RoboDojo tasks~\citep{zhang2026unexpected}.

However, because these frontier foundation models are proprietary, research on RUAs has been limited to using the models as-is, such as harness design~\citep{li2026roboharness,rua2,harness3}.
Without access to their weights, researchers cannot directly explore how further training of these models can advance RUA capabilities.
Moreover, API dependence makes large-scale interaction costly~\citep{openai2026astra} and restricts model-level optimization, such as efficient caching, despite inference speed being a major bottleneck for RUA~\citep{isola2026robotuse}.

The main challenge in adapting open-weight models for robot use is the capability gap between open-weight and closed-weight foundation models. Robot use demands strong spatial understanding and agentic capabilities, which existing open-weight models often lack. Consequently, prior work has focused on ad-hoc improvements, including leveraging privileged information such as ground-truth simulator states or oracle task-success feedback~\citep{fu2026cap,zhang2026harnessvla}, or incorporating task-specific skills~\citep{lu2026aspire}.
% Instead of immediately pursuing such improvements, we argue that establishing a strong foundation for systematically understanding and maximizing the capabilities of existing open-weight models is an essential first step toward more effective robot-specific training.
Before attributing these limitations to the models themselves and pursuing ad-hoc improvements, we argue that we must first establish a generalizable and effective robot-use harness to systematically understand and maximize their existing capabilities, laying the foundation for future RUA training.
% or task-specific skills.
%While effective, these strategies rely on additional training or human-crafted guidance.
%some harnesses utilized previliged informations to fill the gap, such as precise simulator-provided depth or task-verifier feedback to localize objects or assess task progress, which may be unavailable in realistic deployment scenarios. 
%For example, some harnesses rely on precise simulator-provided depth or task-verifier feedback to localize objects or assess task progress, which may be unavailable in realistic deployment scenarios. 

% In this paper, we take a first step toward democratizing robot-use agents for research with open-weight models. We propose \method, a unified harness that enables open-weight models to compete with closed-weight foundation models. Our key insight is that open-weight models can close this gap when equipped with camera parameters, full robot geometry, sensor feedback, and reusable tools and skills, without relying on privileged simulator information. Moreover, we unify the interface and modularize the system components, making \method easy to apply to new benchmarks and models. \method also supports parallel, pipelined execution for efficient batched experiments.

% In this paper, we take a first step toward democratizing robot-use agent research with open-weight models. We propose \method, a harness effective for not only closed-weight foundation models, but also gets the max out of open-weight models. 
% that enables open-weight models to compete with closed-weight foundation models. 
In this paper, we take a first step toward democratizing robot-use agent research by laying this foundation for both competitive open-weight agents and future robot-specific training. We introduce \method, a general-purpose RUA harness. It provides \textbf{Unified} tool interfaces and feedback formats across models, embodiments, tasks, and environments, supporting joint training on simulation and real-world interaction data.
For future large-scale training, we emphasize \textbf{Efficiency} by overlapping model reasoning with tool execution to accelerate interaction, and \textbf{Modularity} through interchangeable agent runtimes, tools, and robot backends.
Together, these designs provide a scalable, extensible foundation for RUA research.

The proposed \method enables competitive manipulation performance with open-weight models, including Qwen and DeepSeek, in both simulation and physical environments.
% While some prior systems obtain performance gains through privileged inputs, we follow our \textbf{Realistic} principle and exclude such inputs.
% \textbf{So, how can we unlock stronger robot-use performance from open models?} \textbf{Our experiments suggest that well-structured information and reliable tool interaction provide effective support for robot use.} For example, camera calibration and robot descriptions support spatial reasoning, while dependable tool execution and informative feedback help agents carry out actions, assess outcomes, and recover from failures.
A key finding is that well-structured information, such as camera calibration and robot descriptions, together with reliable tool interfaces that provide informative feedback, are \textbf{sufficient} for effective robot use. As a result, \textbf{without privileged inputs}, Qwen3.8-Flash-Next with \method achieves a Score of \textbf{31.39}, surpassing GPT-6 Astra's result (\textbf{28.97}) on RoboDojo~\citep{chen2026robodojo} from the RoboDojo team~\citep{zhang2026unexpected}. To the best of our knowledge, this is the first system to outperform closed-weight foundation models on RoboDojo using open-weight models, without privileged informations or distillation from those closed-weight models.
In real-world experiments on 19 tabletop manipulation tasks, Qwen3.8-Flash-Next with \method achieves a Score of \textbf{53.50}, outperforming Dream-1 (\textbf{43.50}; \citealp{dreamlabs2026dream1}) and matching fine-tuned $\pi_{0.5}$ (\textbf{47.75}; \citealp{pi2025pi05}).

Having established the competitiveness of open-weight RUAs, we further compare models across generations, families, and sizes to investigate the capabilities underlying robot use.
Trajectory analysis reveals four common failure patterns involving visual understanding, spatial grounding, geometric coding, and embodied reasoning.
Based on these findings, we posit that current robot-use capability emerges from a combination of these general-purpose capabilities and identify them as key areas for improvement in future RUA training.
Finally, we provide an analysis of the effect of privileged information, cost, and execution efficiency.
% via (Sections~\ref{sec:privileged-information} and~\ref{sec:cost-speed}).

\begin{figure}[t]
\centering
\includegraphics[width=\linewidth]{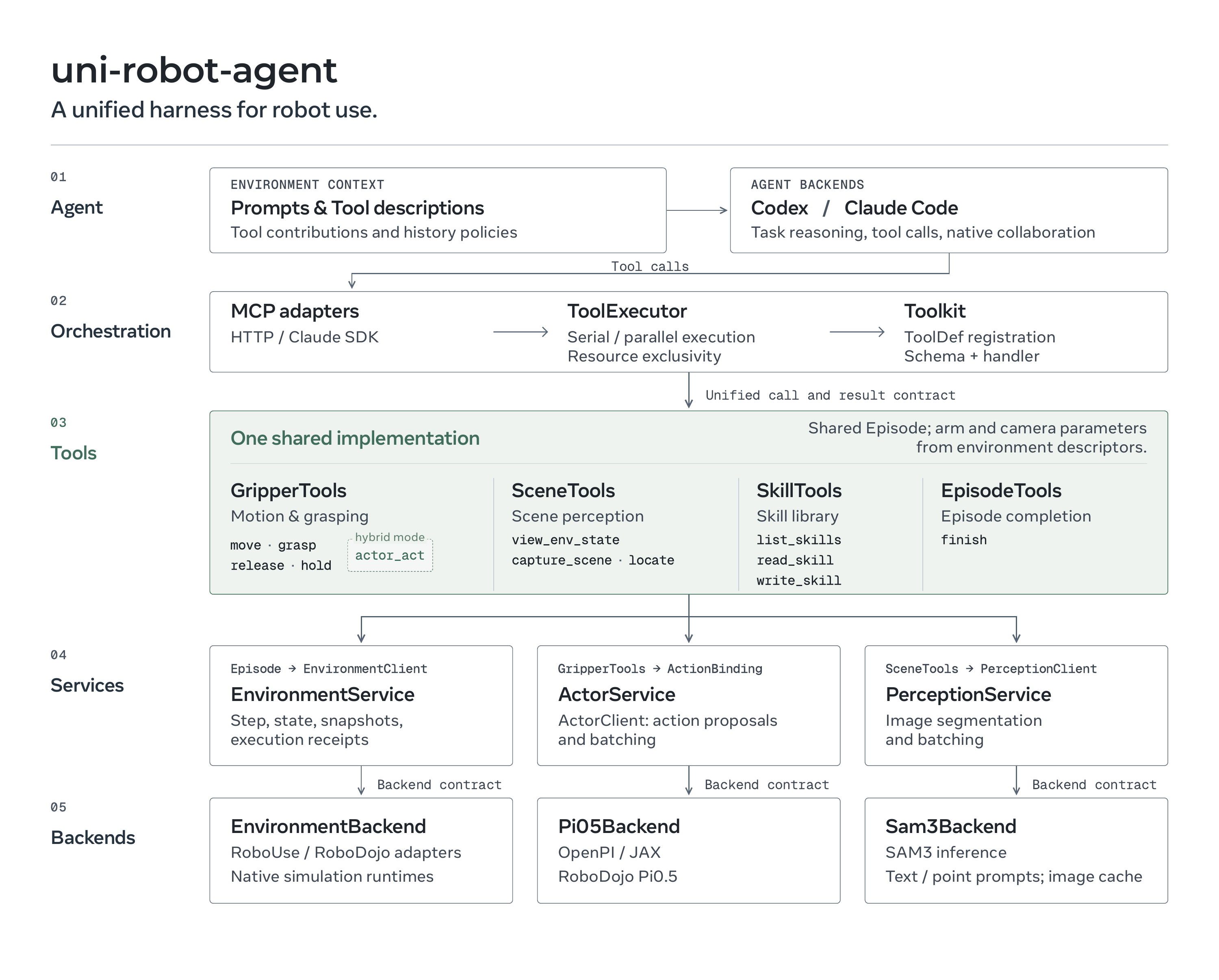}
\caption{\textbf{Five logical layers of \method.} Requests flow from agent runtimes through orchestration and shared tools to domain services and native backends. Context policies and concrete backends are replaceable components; arrows summarize logical interactions rather than individual network connections.}
\label{fig:architecture}
\end{figure}

\section{\method: A Unified and Modular Harness}
\label{sec:harness}

In this section, we introduce the key mechanisms of \method. We describe how the harness connects diverse agents, robot environments, and embodiments through shared tools and supports interaction based on information only available in real-world deployments.

\subsection{System Overview}
\label{sec:system-overview}

\method connects different agent runtimes to a shared toolkit for simulation and physical robots. The agent selects operations using task instructions and observations, while the harness executes them and returns feedback. We organize this interaction around four principles:

\begin{itemize}
\item \textbf{Unified.} Shared tools give operations and feedback a common meaning across environments, with adapters handling native APIs (Section~\ref{sec:unified-interface}). The resulting interaction records can also be combined for subsequent training.
\item \textbf{Efficient.} Reasoning should overlap physical execution, while operations that share a resource retain their required order (Section~\ref{sec:efficient-execution}).
\item \textbf{Realistic.} Agents and tools should rely on information available in physical deployments, without simulator-provided depth or task-verifier feedback (Section~\ref{sec:realistic-inputs}).
\item \textbf{Modular.} Explicit interfaces should allow researchers to replace runtimes, tools, and perception models independently. We introduce this architecture below.
\end{itemize}

These principles guide how we organize the system, but reliable interaction also requires care in how observations, actions, and feedback are handled. We first describe the architecture, then examine these practical design choices in Section~\ref{sec:key-design-choices}.

\paragraph{Modular Architecture.}
A tool call passes through five layers with explicit interfaces (Figure~\ref{fig:architecture}). The \textbf{agent layer} manages model interaction and context, passing tool requests to the \textbf{orchestration layer} for scheduling and dispatch. The \textbf{tool layer} implements shared robot operations and accesses environment, perception, and action-model capabilities through the \textbf{service layer}, which manages their runtime resources. The \textbf{backend layer} supplies the concrete simulator, robot, and model implementations used by these services. Results return through these interfaces to the agent, completing the interaction.

\Needspace{5\baselineskip}
\subsection{Unified Interaction Interface}
\label{sec:unified-interface}

\noindent\begin{minipage}[t]{0.49\textwidth}
\vspace{0pt}
Moving a gripper to a target pose should mean the same thing in simulation and on a physical robot, even when their control APIs differ. We give agents a shared set of tools for motion, perception, skills, and episode control, with common arguments, units, and feedback formats. Environment descriptions identify the available arms, cameras, and optional capabilities. This lets the agent use familiar operations while accounting for the robot and sensors available in each environment.
\end{minipage}\hfill
\begin{minipage}[t]{0.48\textwidth}
\vspace{0pt}
\centering
\captionsetup{font=normalsize,skip=5pt}
\includegraphics[width=\linewidth]{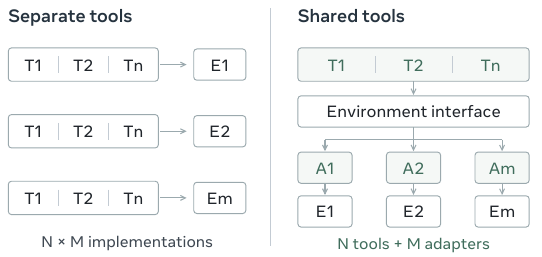}
\captionof{figure}{\textbf{Tool reuse across environments.} Shared tools avoid $N\times M$ implementations. $T$: tool; $A$: adapter; $E$: environment.}
\label{fig:tool-reuse}
\end{minipage}
\par

To make these operations work across environments, shared tools access observations, robot state, and primitive actions through a common interface implemented by environment adapters.
For example, the shared \texttt{move} tool generates intermediate targets and monitors progress; the adapter translates those targets into native commands and returns the resulting state.
Adding an environment therefore requires an adapter and its configuration rather than a separate implementation of every tool.
As Figure~\ref{fig:tool-reuse} illustrates, $N$ tools can share their logic across $M$ adapters, avoiding $N\times M$ tool--environment implementations.
The common request and feedback formats also make the experience collected across environments easier to combine for subsequent training.

\subsection{Efficient Execution}
\label{sec:efficient-execution}

The orchestration layer determines how agent interaction proceeds while tools are running. Waiting for every movement to finish delays reasoning that could proceed independently, yet successive movements may still need to execute in order. We combine background tool calls with resource-aware scheduling to satisfy both requirements, forming a pipeline between agent reasoning and tool execution.

\paragraph{Two Levels of Concurrency.}
The agent need not wait for every tool, but its next call may still need to wait for a resource:
\begin{itemize}[topsep=3pt,itemsep=2pt,parsep=0pt]
\item \textbf{Between agent and tools.} A \texttt{background} call returns a job handle, allowing the agent to continue independent reasoning and retrieve the result later; an \texttt{immediate} call waits for completion. Decisions that depend on the result must still wait.
\item \textbf{Among tool calls.} Calls sharing an exclusive resource execute in submission order; parallel calls run outside that queue. Two \texttt{move} calls on the same resource therefore run sequentially, while reading an existing snapshot can overlap movement.
\end{itemize}

\par\addvspace{\baselineskip}
\noindent\begin{minipage}[t]{0.49\textwidth}
\vspace{0pt}
\textbf{Two-Stage Pipelining.}\quad These mechanisms enable a two-stage pipeline comprising agent reasoning and tool execution (Figure~\ref{fig:agent-tool-pipeline}). The agent can submit successive calls while previously submitted calls execute in resource order. Reasoning and execution thus overlap even when the tools themselves execute sequentially. When observations or execution feedback require a change of plan, the agent can use \texttt{flush\_queue} to cancel calls that have not started for a specified resource. Where supported, a single model response can submit multiple tool requests, and a batched wait can retrieve multiple results.
\end{minipage}\hfill
\begin{minipage}[t]{0.48\textwidth}
\vspace{0pt}
\centering
\captionsetup{font=normalsize,skip=5pt}
\includegraphics[width=\linewidth]{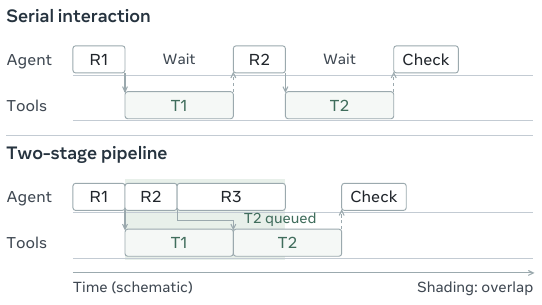}
\captionof{figure}{\textbf{Serial interaction and pipelining.} Reasoning ($R$) overlaps sequential tools ($T$); $T_2$ queues while $T_1$ runs. Solid/dashed arrows show submissions/results; shading marks overlap. Timings are schematic.}
\label{fig:agent-tool-pipeline}
\end{minipage}
\par

\subsection{Interaction Without Privileged Inputs}
\label{sec:realistic-inputs}

Efficient execution alone does not ensure that simulation performance carries over to physical robots: the information supporting each action also matters. Existing systems provide depth-based geometry and completion checks~\citep{li2026roboharness}, or construct object models to maintain a virtual scene~\citep{wang2026spatialharness}. In physical settings, depth measurements can vary or be missing even in a static scene, and simulator task-verifier feedback is unavailable. These differences motivate testing whether agents can infer spatial relationships and assess progress from more limited observations.

% \begin{figure}[t]
% \centering
% \captionsetup{subrefformat=parens}
% \captionsetup[subfigure]{position=top,labelformat=parens,labelsep=space,font=footnotesize,labelfont=bf,textfont=bf,skip=4pt,justification=raggedright,singlelinecheck=false}
% \begin{subfigure}[t]{0.360\linewidth}
% \phantomsection\phantomcaption
% \label{fig:depth-fixed-pixel}
% \includegraphics[width=\linewidth]{figs/realistic-inputs/depth-fixed-pixel.pdf}
% \end{subfigure}\hfill
% \begin{subfigure}[t]{0.345\linewidth}
% \phantomsection\phantomcaption
% \label{fig:spatial-variability}
% \includegraphics[width=\linewidth]{figs/realistic-inputs/spatial-variability.pdf}
% \end{subfigure}\hfill
% \begin{subfigure}[t]{0.275\linewidth}
% \phantomsection\phantomcaption
% \label{fig:task-feedback}
% \includegraphics[width=\linewidth]{figs/realistic-inputs/task-feedback.pdf}
% \end{subfigure}
% \caption{\textbf{Information conditions for robot interaction.} \subref{fig:depth-fixed-pixel} Fixed-pixel depth versus a schematic constant reference at the measured median. \subref{fig:spatial-variability} Depth variability and missingness in a static scene (1,000 frames, 33.3\,s). \subref{fig:task-feedback} Simulator completion feedback, unavailable on physical robots. Our verifier is used only for evaluation.}
% \label{fig:realistic-inputs}
% \end{figure}

\subsection{Key Design Choices}
\label{sec:key-design-choices}

Effective robot use requires careful design of how observations reach the agent, how commands are executed, and how outcomes inform subsequent decisions. We introduce six practical design choices spanning observation and execution consistency, coordinated motion control, progress monitoring, actionable feedback, and visual context management.

\begin{figure}[t]
\centering
\captionsetup{subrefformat=parens}
\captionsetup[subfigure]{position=top,labelformat=parens,labelsep=space,font=footnotesize,labelfont=bf,textfont=bf,skip=4pt,justification=raggedright,singlelinecheck=false}
\begin{minipage}[t]{0.318\linewidth}
\vspace{0pt}

\begin{subfigure}[t]{\linewidth}
\phantomsection\phantomcaption
\label{fig:design-snapshot-consistency}
\includegraphics[width=\linewidth]{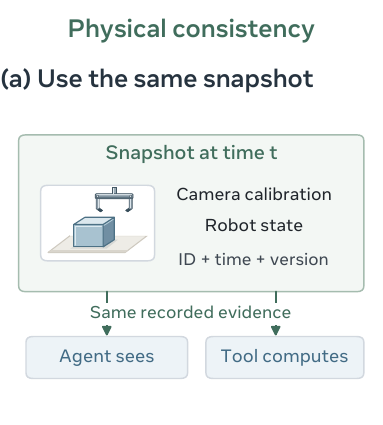}
\end{subfigure}
\par\medskip
\begin{subfigure}[t]{\linewidth}
\phantomsection\phantomcaption
\label{fig:design-uncertain-execution}
\includegraphics[width=\linewidth]{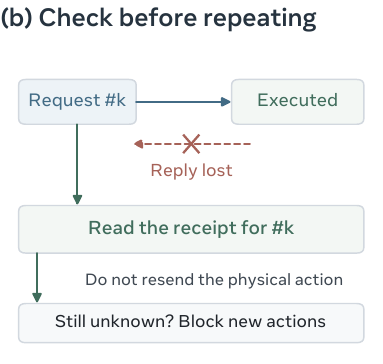}
\end{subfigure}
\end{minipage}\hfill
\begin{minipage}[t]{0.318\linewidth}
\vspace{0pt}

\begin{subfigure}[t]{\linewidth}
\phantomsection\phantomcaption
\label{fig:design-joint-control}
\includegraphics[width=\linewidth]{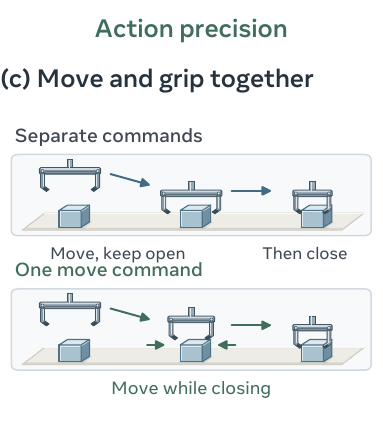}
\end{subfigure}
\par\medskip
\begin{subfigure}[t]{\linewidth}
\phantomsection\phantomcaption
\label{fig:design-monitored-motion}
\includegraphics[width=\linewidth]{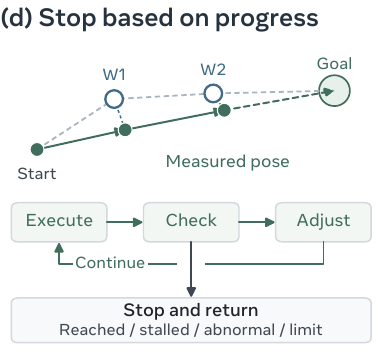}
\end{subfigure}
\end{minipage}\hfill
\begin{minipage}[t]{0.318\linewidth}
\vspace{0pt}

\begin{subfigure}[t]{\linewidth}
\phantomsection\phantomcaption
\label{fig:design-agent-feedback}
\includegraphics[width=\linewidth]{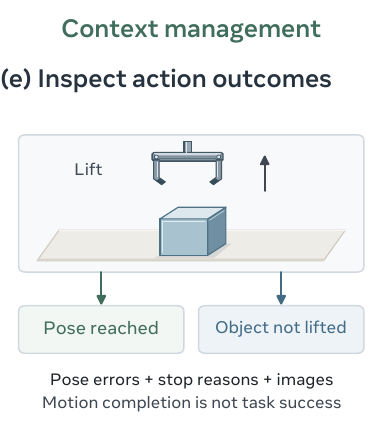}
\end{subfigure}
\par\medskip
\begin{subfigure}[t]{\linewidth}
\phantomsection\phantomcaption
\label{fig:design-visual-context}
\includegraphics[width=0.9\linewidth]{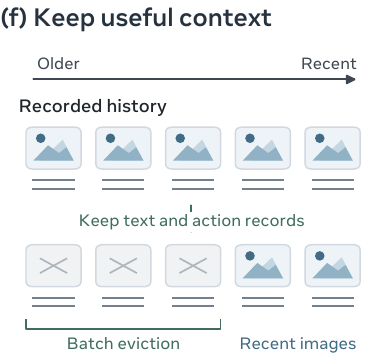}
\end{subfigure}
\end{minipage}
\caption{\textbf{Key design choices.} Each column pairs two designs for physical consistency, action precision, or context management.}
\label{fig:key-design-choices}
\vspace{-4mm}
\end{figure}

\paragraph{Keeping Observations and State Consistent.}
When a robot moves after an image is captured, interpreting a selected pixel using a newer camera pose mixes two physical states. We bind images, camera calibration, and robot state to an immutable snapshot and use that snapshot for localization (Figure~\ref{fig:design-snapshot-consistency}). The agent and tools thus work from the same recorded state.

\paragraph{Preventing Repeated Actions After Communication Failures.}
The agent must also know which actions have occurred: a missing reply may mean lost feedback, so retrying could repeat a physical action. We record request identities and execution status to reject duplicates (Figure~\ref{fig:design-uncertain-execution}). If completion is uncertain, further actions are blocked while the session attempts to recover the execution record; unresolved requests remain explicitly marked as uncertain.

\paragraph{Coordinating Motion and Gripper Control.}
Separate movement and gripper commands prevent the agent from adjusting the opening during a movement. Our \texttt{move} tool accepts a target pose and an optional gripper opening together, allowing the arm and gripper to act simultaneously (Figure~\ref{fig:design-joint-control}).

\paragraph{Detecting When Motion Should Stop.}
During execution, a fixed step count may stop motion too early or waste steps on an unreachable target. Our \texttt{move} tool generates intermediate targets from measured poses and checks progress after each step (Figure~\ref{fig:design-monitored-motion}). This lets the tool adjust motion locally, stop on arrival, and return control when motion stalls or reaches an execution limit. Appendix~\ref{app:motion-execution} gives the algorithm and stopping criteria.

\paragraph{Making Action Outcomes Diagnosable.}
Reaching a target pose does not establish that an object was grasped or placed successfully. We return numerical motion feedback, including pose errors and stopping reasons, alongside images of the outcome (Figure~\ref{fig:design-agent-feedback}). The numbers distinguish a small residual error from rejected or stalled motion; the images show the effect on the object. Together, they help the agent decide whether to refine the motion or change its manipulation strategy.

\paragraph{Managing Growing Visual Histories.}
This feedback accumulates, but old images may lose relevance while task instructions and action records remain useful. Our sliding window replaces old images with text placeholders, preserving recent images and the surrounding text (Figure~\ref{fig:design-visual-context}). We evict images in batches to keep the prompt prefix stable between evictions and support cache reuse. If the context still exceeds budget, we compact the history with the same model that we use for the policy, using the compaction prompt of the codex CLI~\citep{codex}.

%we remove older complete interactions, keeping tool calls with their results and protecting essential instructions and pending calls. 

\section{Can Open Models Use Robots in Simulation and the Real World?}
\label{sec:evaluation}

Effective robot use has largely been the preserve of frontier, closed-source models. In this work, however, we show that open-weight models can also achieve competitive performance in simulation and on physical robots. In particular, on RoboDojo, Qwen3.8-Flash-Next surpasses most fine-tuned VLA and WAM baselines in a zero-shot setting without privileged information, and also the GPT-6 Astra's result reported by the RoboDojo team.
On real robots, a less-explored setting in prior works, open-weight models also achieve competitive performance, showing their real-world applicability.
These results highlight the viability of open-source robot-use agents and the importance of a properly designed harness for supporting these models.

\subsection{Simulation Evaluation}
\label{sec:simulation-results}

\subsubsection{Setup}

We evaluate 42 RoboDojo~\citep{chen2026robodojo} tasks spanning generalization, memory, precision, long-horizon execution, and open-vocabulary instruction following. We report the success rate and the score (incorporating partial success, scaled to 0--100) following the RoboDojo's evaluation protocol. The result is averaged over three seeds with two layouts (six episodes per task, 252 episodes in total).
Appendix~\ref{app:evaluation-budget} examines Score stability across rollout counts.
The 12 generalization tasks pair equally weighted standard and randomized layouts per seed to test robustness to object and scene changes. Following the official protocol, we average the results of each task within the dimension, and average those results over the five dimensions.
We allow up to \textbf{3h} of wall-clock time per episode, as aggregate performance largely plateaued by the third hour in four-hour runs.

\begin{table*}[t]
\centering
\caption{\textbf{Results on RoboDojo.} Each entry reports SR/Score, where success rate (SR) is a percentage and Score is on a 0--100 scale. Our \method configurations exclude task-verifier feedback. Depth-enabled results are shown in gray for reference; our primary results do not use depth. Avg.\ is the unweighted mean of each metric over the five capability dimensions. $^{\dagger}$GAP (Hybrid) incorporates $\pi_{0.5}$.}
% Qwen/DeepSeek source and fixed252 aggregation: outputs/main-nodepth-recheck-20261028/summary.json.
\label{tab:robodojo-main}
\fontsize{8}{10}\selectfont
\setlength{\tabcolsep}{1pt}
\renewcommand{\arraystretch}{1.35}
% Local leaders keep dashed separators compatible with colored rows.
\def\MainTableDashedRule{%
  \noalign{\vskip\aboverulesep}
  \multispan{8}\xleaders\hbox{\rule{1.5pt}{\lightrulewidth}\hspace{2pt}}\hfill\cr
  \noalign{\vskip\belowrulesep}}
\begin{tabularx}{\textwidth}{@{}ll*{6}{>{\centering\arraybackslash}X}@{}}
\toprule
\textbf{Model} & \textbf{Harness} & Gen. & Prec. & Long-Hor. & Memory & Open & \textbf{Avg.}$\uparrow$ \\
\midrule
\multicolumn{8}{@{}c@{}}{VLA \& WAM} \\
\midrule
% Official RoboDojo simulation leaderboard, accessed 2026-10-20:
% https://robodojo-benchmark.com/leaderboard
% Pi-05 rechecked against the official live bundle on 2026-10-25:
% /assets/index-DA5jHulR.js: Memory Score/SR 5.89/4.67, Avg. 11.44/6.93.
% Keep this leaderboard snapshot; the Astra paper used an earlier snapshot.
$\pi_{0.5}$ & -- & 8.17/13.38 & 5.50/12.40 & 14.67/23.54 & 4.67/5.89 & 1.67/1.98 & 6.93/11.44 \\
Hy-Embodied-0.5-VLA & -- & 8.39/11.78 & 8.00/13.81 & 14.92/25.74 & 12.11/13.37 & 0.58/0.65 & 8.80/13.07 \\
OpenWAM-$\alpha$ & -- & 14.83/20.71 & 9.25/18.45 & 25.33/34.93 & 9.11/10.41 & 1.08/1.41 & 11.92/17.18 \\
Xiaomi-Robotics-1 & -- & 17.00/23.54 & 18.83/26.69 & 23.67/38.39 & 6.56/7.81 & 3.58/3.94 & 13.93/20.07 \\
\midrule
\multicolumn{8}{@{}c@{}}{Agent} \\
\midrule
GPT-6 Astra & RoboProbe & 30.50/33.36 & 4.00/12.65 & 8.25/21.45 & 38.67/43.04 & 31.00/34.36 & 22.48/28.97 \\
\rowcolor{paperAccentSoft}
GPT-6 Astra & \textcolor{paperAccent}{\textbf{\method}} & 31.94/37.15 & 20.83/28.75 & 39.58/50.31 & 80.56/80.56 & 60.42/62.71 & 46.67/51.90 \\
\textcolor{gray}{GPT-6 Astra} & \textcolor{gray}{\method{} + depth} & \textcolor{gray}{43.06/47.85} & \textcolor{gray}{29.17/36.04} & \textcolor{gray}{52.08/60.31} & \textcolor{gray}{75.00/75.42} & \textcolor{gray}{62.50/64.17} & \textcolor{gray}{52.36/56.76} \\
\MainTableDashedRule
Qwen3.8-Flash-Next & RoboProbe & 0.00/0.35 & 0.00/0.83 & 0.00/3.75 & 0.00/0.00 & 4.17/4.58 & 0.83/1.90 \\
Qwen3.8-Flash-Next & RPent & 5.56/9.24 & 4.17/8.02 & 4.17/12.40 & 5.56/5.56 & 2.08/3.54 & 4.31/7.75 \\
Qwen3.8-Flash-Next & GAP (Hybrid)$^{\dagger}$ & 13.89/19.31 & 8.33/16.46 & 18.75/28.85 & 16.67/16.67 & 22.92/23.75 & 16.11/21.01 \\
\rowcolor{paperAccentSoft}
DeepSeek-V4.1-Flash & \textcolor{paperAccent}{\textbf{\method}} & 16.67/20.35 & 6.25/16.25 & 4.17/14.79 & 36.11/38.33 & 43.75/45.83 & 21.39/27.11 \\
\textcolor{gray}{DeepSeek-V4.1-Flash} & \textcolor{gray}{\method{} + depth} & \textcolor{gray}{18.06/21.88} & \textcolor{gray}{12.50/19.79} & \textcolor{gray}{10.42/25.31} & \textcolor{gray}{41.67/45.56} & \textcolor{gray}{37.50/38.54} & \textcolor{gray}{24.03/30.22} \\
\rowcolor{paperAccentSoft}
Qwen3.8-Flash-Next & \textcolor{paperAccent}{\textbf{\method}} & \textbf{18.06/21.60} & \textbf{2.08/10.21} & \textbf{6.25/23.75} & \textbf{52.78/54.72} & \textbf{45.83/46.67} & \textbf{25.00/31.39} \\
\textcolor{gray}{Qwen3.8-Flash-Next} & \textcolor{gray}{\method{} + depth} & \textcolor{gray}{22.22/25.28} & \textcolor{gray}{18.75/26.35} & \textcolor{gray}{14.58/29.69} & \textcolor{gray}{52.78/56.25} & \textcolor{gray}{39.58/41.88} & \textcolor{gray}{29.58/35.89} \\
\bottomrule
\end{tabularx}
\vspace{5pt}
\end{table*}

\textbf{No Privileged Inputs.}
%Unlike prior works~\citep{harness1,spatialharness,sear}, 
Our simulation uses the inputs of realistic robotics applications, providing only RGB images and camera calibration, robot proprioception and descriptions, and task instructions. We do not allow access to ground-truth depth from simulators, CAD assets of the object, hidden scene annotations, task-verifier feedback, and external web access. 
%The same restrictions apply to tool computations: localization must derive coordinates from permitted observations, for example through multi-view triangulation, which agents may also implement in code. 
We also manually check that the models did not made any invalid access to the prohibited information throughout the experiment. Within these input constraints, agents still need a clear description of the task. Some RoboDojo instructions omit or ambiguously describe task requirements: the button-pressing instruction does not clearly specify when each red button's press count should be confirmed with the blue button, while the tower-building instruction leaves the arrangement of wooden blocks and boards unspecified. We therefore add brief clarifications of these task rules, without providing scoring rubrics or verifier logic.

To contextualize open-weight model performance, we compare against official Astra--RoboProbe~\citep{zhang2026unexpected} results and representative VLA and WAM baselines from the RoboDojo leaderboard~\citep{robodojo2026leaderboard}. To compare the contribution of the harness, we compare Qwen3.8-Flash-Next across \method, GAP (Hybrid)~\citep{su2026astra}, RoboProbe, and RPent~\citep{zhang2026harnessvla}. Baseline configurations are detailed in Appendix~\ref{app:reproducibility-settings}.

\subsubsection{Results and Analysis}

As shown in Table~\ref{tab:robodojo-main}, Qwen3.8-Flash-Next with \method achieves \textbf{25.00\% SR and 31.39 Score}, surpassing listed VLA/WAM baselines and GPT-6 Astra with RoboProbe (22.48\% SR, 28.97 Score), despite being an open-weight model. DeepSeek-V4.1-Flash also outperforms the VLA/WAM baselines (\textbf{21.39\% SR, 27.11 Score}). Moreover, switching to other harnesses results in a substantially degrades the performance (loss of $8.89\%$ SR and $10.38$ Score on the strongest baseline GAP).
Together, these results demonstrate competitive robot use across two open-weight families and the importance of our harness on this performance.

%Moreover, switching to GAP (Hybrid), the strongest baseline, results in a substantial degradation (loss of $8.89\%$ SR and $10.38$ Score). Moreover, Qwen with \method, despite being open-weight, outperforms GPT-6 Astra with RoboProbe (22.48\% SR, 28.97 Score). When switched to \method, GPT-6 Astra reaches 46.67\% SR and 51.90 Score.

%Looking at the breakdowns in Table~\ref{tab:robodojo-main}, 

Both open-weight or closed-source RUAs tend to achieve higher scores on memory and open-vocabulary tasks than the VLA and WAM baselines. On long-horizon tasks, however, Qwen and DeepSeek perform worse than Xiaomi-Robotics-1 and OpenWAM-$\alpha$. We note that GPT-6 Astra equipped with \method performs the best, showing that the design choices of \method can also benefit closed-weight models too. %Despite these gains, both open-weight models score below GPT-6 Astra across all five dimensions, with particularly large gaps in memory and long-horizon tasks.

\subsection{Real-World Evaluation}
\label{sec:real-world}

\subsubsection{Setup}

We design an in-house benchmark with 19 tasks across five equally weighted dimensions: spatial understanding, object generalization, instruction following, compositional generalization, and unseen tasks. %, inspired from $\pi_{0.5}$'s evaluation categories.
We evaluate Qwen3.8-Flash-Next and GPT-6 Astra with \method, alongside Dream-1~\citep{dreamlabs2026dream1} in a zero-shot setting, while $\pi_{0.5}$~\citep{pi2025pi05} is fine-tuned on 10 hours of demonstrations per task. All methods use the same dual-arm platform. We note that our agents additionally receive depth from Intel RealSense D405 cameras, while $\pi_{0.5}$ and Dream-1 use RGB only. 
%This input difference limits direct comparison, a limitation of this work, as our primary aim is to demonstrate real-world feasibility with RUAs.
Figure~\ref{fig:real-world-tasks} shows details of representative real-world tasks.
Appendix~\ref{app:real-eval-benchmark} provides the task catalogue, scoring protocol, hardware, and baseline details.

\par\addvspace{8pt}
\noindent\begin{minipage}{\linewidth}
  \centering
  \includegraphics[width=\linewidth]{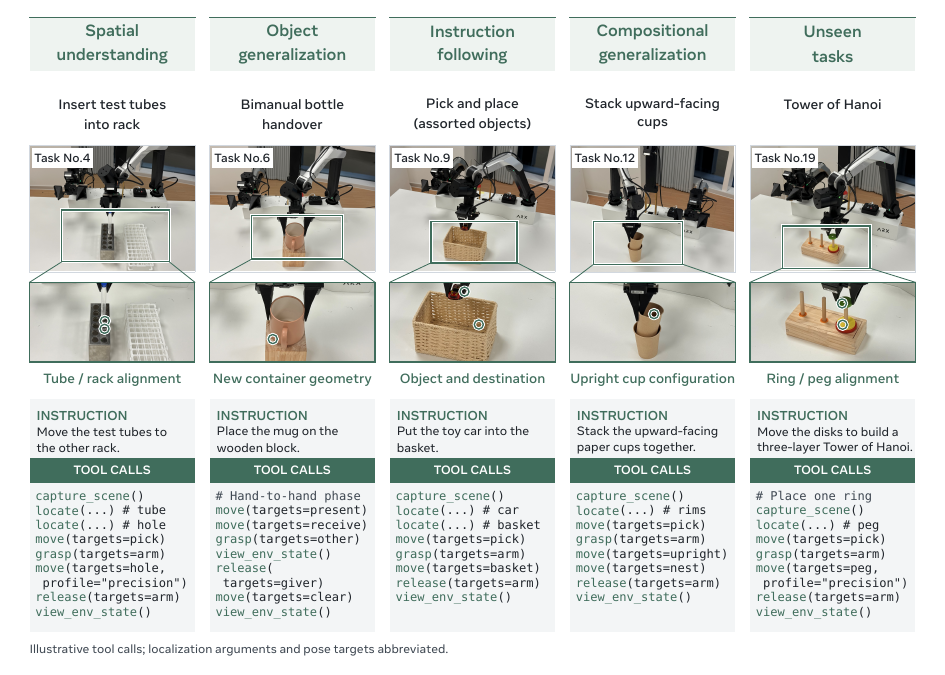}
  \captionof{figure}{\textbf{Representative real-world tasks across five dimensions.} Scene photographs and close-ups show task details, with example instructions and illustrative tool calls.}
  \label{fig:real-world-tasks}
\end{minipage}
\par\addvspace{8pt}

\subsubsection{Results and Analysis}

As shown in Table~\ref{tab:real-eval-summary}, Qwen3.8-Flash-Next with \method matches the performance of fine-tuned $\pi_{0.5}$, and outperforms zero-shot Dream-1 in real world experiments. When applied to GPT-6 Astra, it outperforms other baselines with a large margin. Generally, RUAs perform better on compositional tasks and unseen tasks, while VLA/WAMs performing well on object generalization tasks. These results show that \method can be also applied to real world applications. Please refer to Table~\ref{tab:real-eval-per-task} for detailed per-task results.

\begin{table*}[!t]
\centering
\caption{\textbf{Results on real-world evaluation.} Each capability is scored out of 20; Total is out of 100. Per-task results and scoring details are provided in Appendix~\ref{app:real-eval-benchmark}.}
\label{tab:real-eval-summary}
\small
\setlength{\tabcolsep}{4pt}
\renewcommand{\arraystretch}{1.25}
\begin{tabularx}{\textwidth}{@{}
  >{\raggedright\arraybackslash}p{0.3\textwidth}
  >{\hsize=0.7\hsize\linewidth=\hsize\centering\arraybackslash}X
  >{\hsize=0.7\hsize\linewidth=\hsize\centering\arraybackslash}X
  >{\hsize=1.0\hsize\linewidth=\hsize\centering\arraybackslash}X
  >{\hsize=1.6\hsize\linewidth=\hsize\centering\arraybackslash}X
@{}}
\toprule
 & \textbf{VLA} & \textbf{WAM}
 & \multicolumn{2}{c}{\textcolor{paperAccent}{\textbf{\method}}} \\
\cmidrule(lr){2-2}\cmidrule(lr){3-3}\cmidrule(l){4-5}
\textbf{Capability}
 & $\pi_{0.5}$ & Dream-1 & GPT-6 Astra
 & \mbox{Qwen3.8-Flash-Next} \\
\midrule
Spatial understanding         & 11.25 &  6.50 & 17.00 &  9.00 \\
Object generalization         & 12.50 & 13.50 & 17.50 & 10.00 \\
Instruction following         & 14.00 &  9.50 & 18.00 & 13.50 \\
Compositional generalization  &  6.00 & 10.00 & 20.00 & 13.00 \\
Unseen tasks                  &  4.00 &  4.00 &  8.00 &  8.00 \\
\midrule
\rowcolor{paperAccentSoft}
\textbf{Total}$\uparrow$       & 47.75 & 43.50 & \textbf{80.50} & 53.50 \\
\bottomrule
\end{tabularx}
\end{table*}

\section{What Capabilities Enable Robot-Use Agents?}
\label{sec:model-capabilities}

The results in Section~\ref{sec:evaluation} establish the feasibility of robot use with open-weight models, but leave an important question: what enables these models to work? To investigate this, we examine three comparisons. First, we revisit earlier model generations, including Qwen3.5--3.7 and GPT-5.6. Second, we compare recent models from different families, including Qwen, DeepSeek, GLM, and GPT. Third, we examine Qwen models of different sizes within the same generation.

% \subsection{Setup}
% \label{sec:model-capabilities-setup}

We evaluate the additional models on RoboDojo's 42 tasks and report SR and Score using the same aggregation as Section~\ref{sec:evaluation}. We reuse the Section~\ref{sec:evaluation} results for Qwen3.8-Flash-Next, DeepSeek-V4.1-Flash, and GPT-6 Astra. The additional evaluations use one seed and one layout.
%We note that these additional evaluations are exploratory case studies rather than controlled statistical comparisons. %We therefore use the results to identify patterns for further investigation and do not claim statistical significance.

\subsection{Performance Across Generations, Families, and Model Sizes}
\label{sec:model-capabilities-results}

\begin{figure}[!htbp]
\centering
\includegraphics[width=\linewidth]{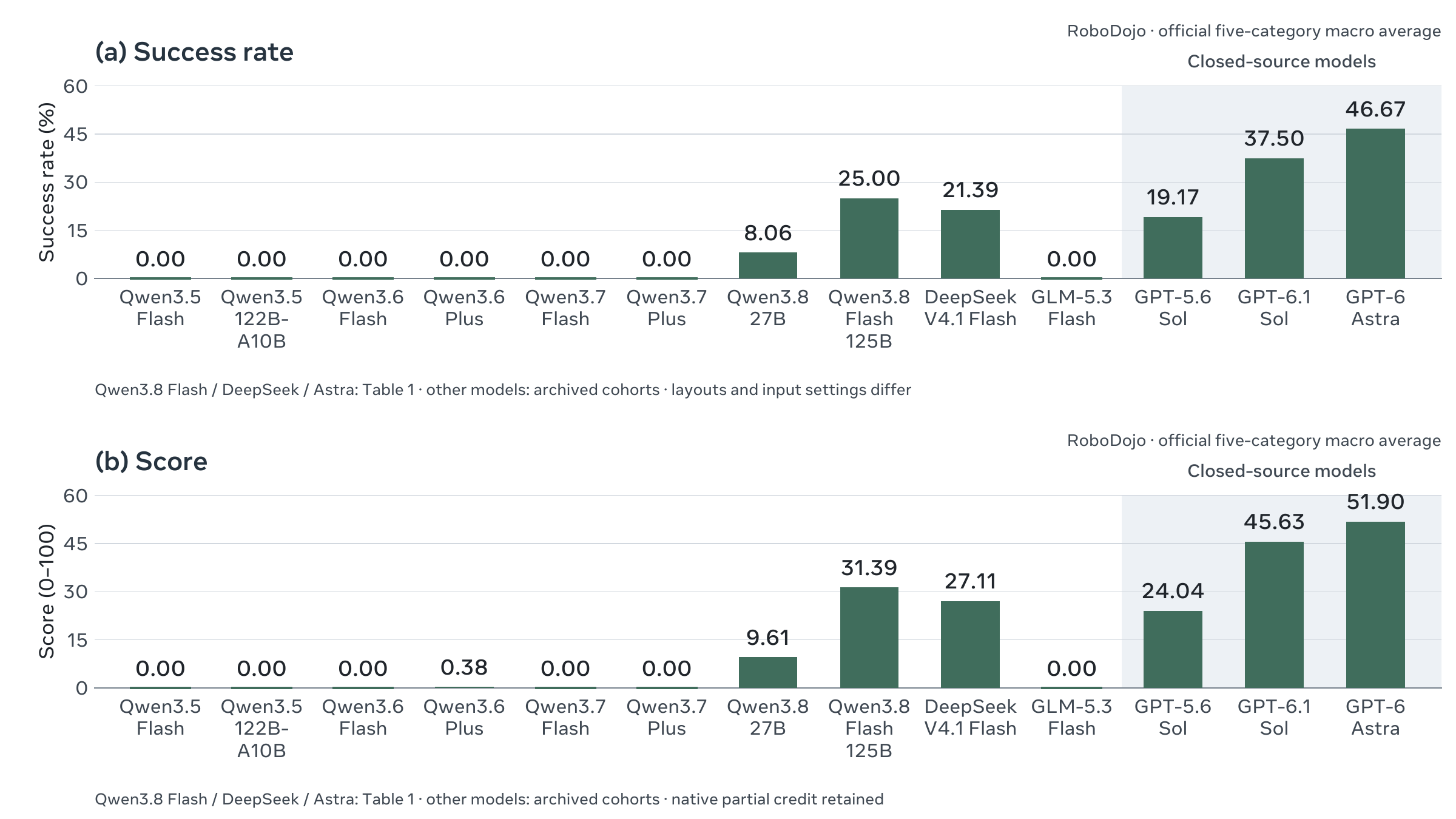}
\caption{\textbf{Robot-use performance across model generations, families, and sizes.}
Success rate (top) and Score (bottom) on RoboDojo with \method.
Robot use is neither exclusive to frontier models nor unique to the latest generation: GPT-5.6 Sol already achieves nontrivial performance, while Qwen3.8-27B succeeds where larger, earlier Qwen models fail. Shading indicates closed-source models.}
\label{fig:thirteen-model-results}
\vspace{-4mm}
\end{figure}

\paragraph{Robot use predates the latest model generation.}
Figure~\ref{fig:thirteen-model-results} shows that GPT-5.6 Sol already achieves a nontrivial \textbf{19.17\% SR and 24.04 Score}, while GPT-6.1 Sol reaches \textbf{37.50\% SR and 45.63 Score}. Thus, the latest generation substantially improves robot-use performance, but these capabilities were already present in earlier models.

\paragraph{The pattern differs across model families.}
However, among the models evaluated, only the GPT series demonstrates successful robot use in generations preceding the latest one. For Qwen, verified task success is observed only in the 3.8 generation: Qwen3.5--3.7 Flash, Qwen3.6--3.7 Plus, and Qwen3.5-122B record zero SR, whereas Qwen3.8-27B and Qwen3.8-Flash-Next achieve \textbf{8.06\%} and \textbf{25.00\%}, respectively. More importantly, a recent model generation does not by itself guarantee task success: GLM-5.3 Flash shows $0\%$ success rate as older Qwen models.

\paragraph{Model size matters less than training distribution.}
Within Qwen3.8, Flash-Next (125B) achieves higher SR and Score than the 27B model. Across generations, however, Qwen3.8-27B completes tasks while all earlier Qwen models evaluated, including Qwen3.5-122B, Qwen3.7 Plus ($>150$B parameters), and Qwen3.6 Plus ($>150$B parameters), record zero SR. We posit that training distributions differ substantially across generations and that these differences matter more for robot use than model size alone.

Taken together, these results suggest that robot use is neither exclusive to frontier models nor a capability that emerged uniformly with the latest generation. Earlier GPT models already achieve nontrivial performance, while Qwen3.8-27B succeeds where larger, earlier Qwen models fail. Rather than model size or recency alone, these contrasting trajectories point to differences in the underlying capabilities (i.e. training distribution) developed across model families and generations. We next examine failure modes to identify which capabilities distinguish successful robot use from failure.

\begin{figure}[t]
\centering
\includegraphics[width=0.72\linewidth]{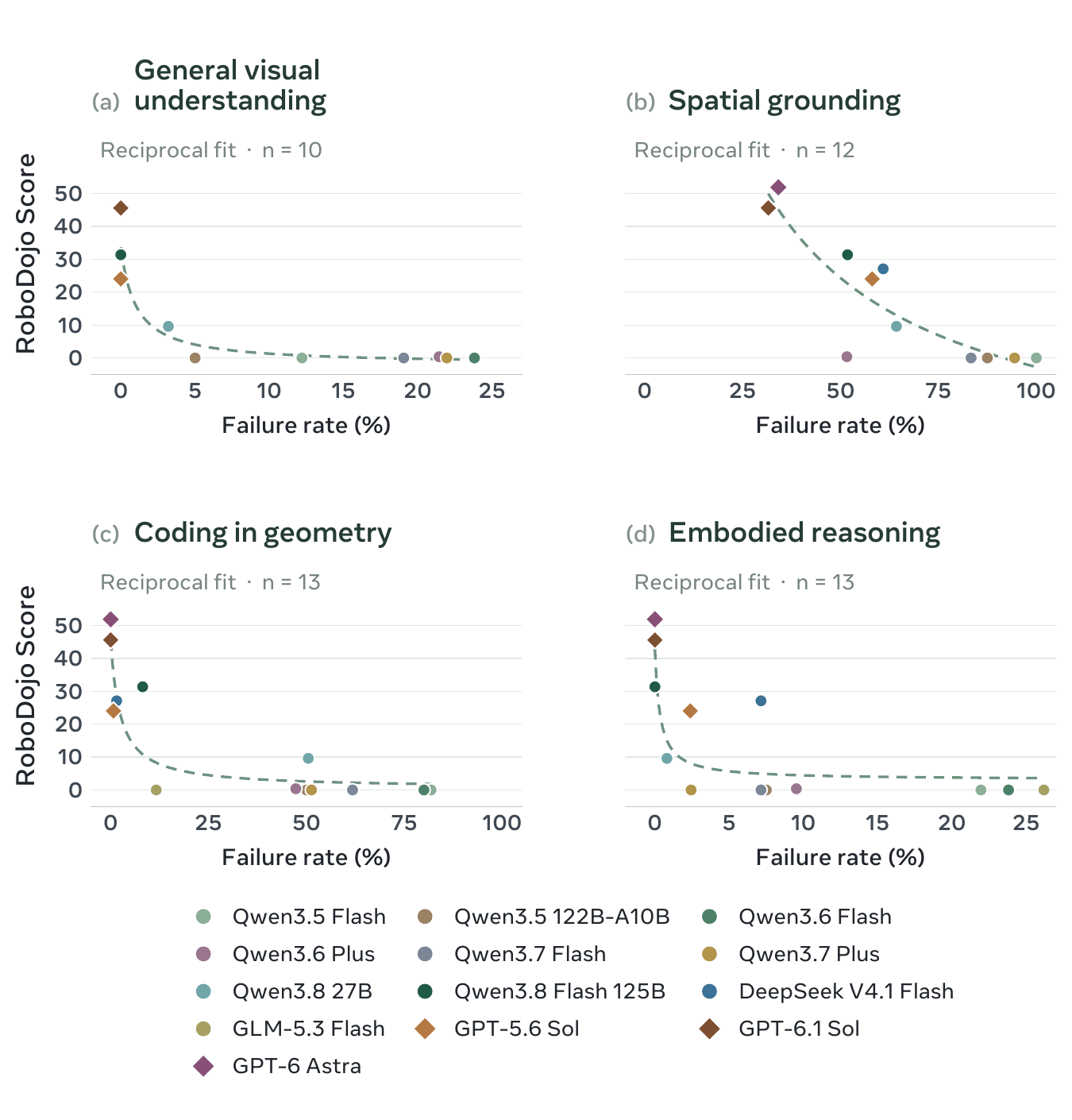}
\caption{\textbf{Four capabilities associated with effective robot use.}
Correlation between each model's RoboDojo Score and its errors in
(a) general visual understanding, (b) spatial grounding,
(c) coding in geometry, and (d) embodied reasoning.
Lower error rates across these dimensions are associated with higher task performance, suggesting that effective robot use draws on a combination of these capabilities.}
\label{fig:four-failure-modes}
\vspace{-4mm}
\end{figure}

\subsection{Four Major Failure Modes}
\label{sec:four-failure-modes}

Our trajectory analysis identifies four major failure types: \emph{general visual understanding}, such as misinterpreting objects or action outcomes; \emph{fine-grained pointing and spatial grounding}, such as selecting inconsistent corresponding points across views; \emph{localization and pose computation code}, involving errors in geometric calculations, coordinate transformations, or orientation construction; and \emph{embodied reasoning}, such as proposing physically inappropriate grasp poses. Figure~\ref{fig:four-failure-modes} summarizes these four failure types across thirteen models.

\paragraph{General visual understanding.}
Perhaps the most straightforward reason for failures in earlier Qwen Flash and Plus models is basic visual misunderstanding. Such errors, including mistaking zero for nine or inverted cups for upright ones, occur in \textbf{12.2--23.8\%} of trials. In contrast, these kinds of errors occur in \textbf{3.2\%} of Qwen3.8-27B trials and none of the trials from more performant models such as Qwen3.8-Flash or GPT-5.6 Sol. These observations suggest that general visual understanding is closely associated with its performance.

%This pattern suggests that reliable general visual understanding is closely associated with eventual task success.

\paragraph{Spatial grounding.}
In addition to coarse scene understanding, we find that models struggle to pinpoint the same physical feature across camera views. Selecting inconsistent pixels for the same feature leads to large reprojection errors, causing the triangulation tool to reject the result. We therefore use the fraction of calls rejected for this reason to measure difficulty in pinpointing corresponding features. Task performance tends to be negatively associated with this failure rate: Qwen3.8-Flash and DeepSeek V4.1 Flash have lower failure rates than models with zero task success. GPT-6.1 Sol achieves the lowest spatial-grounding failure rate among the evaluated models (31.5\%), but this substantial remaining error rate highlights precise spatial grounding as a common challenge for further improvement (Figure~\ref{fig:four-failure-modes}b).

\begin{figure}[t]
  \centering
  \includegraphics[width=0.72\linewidth]{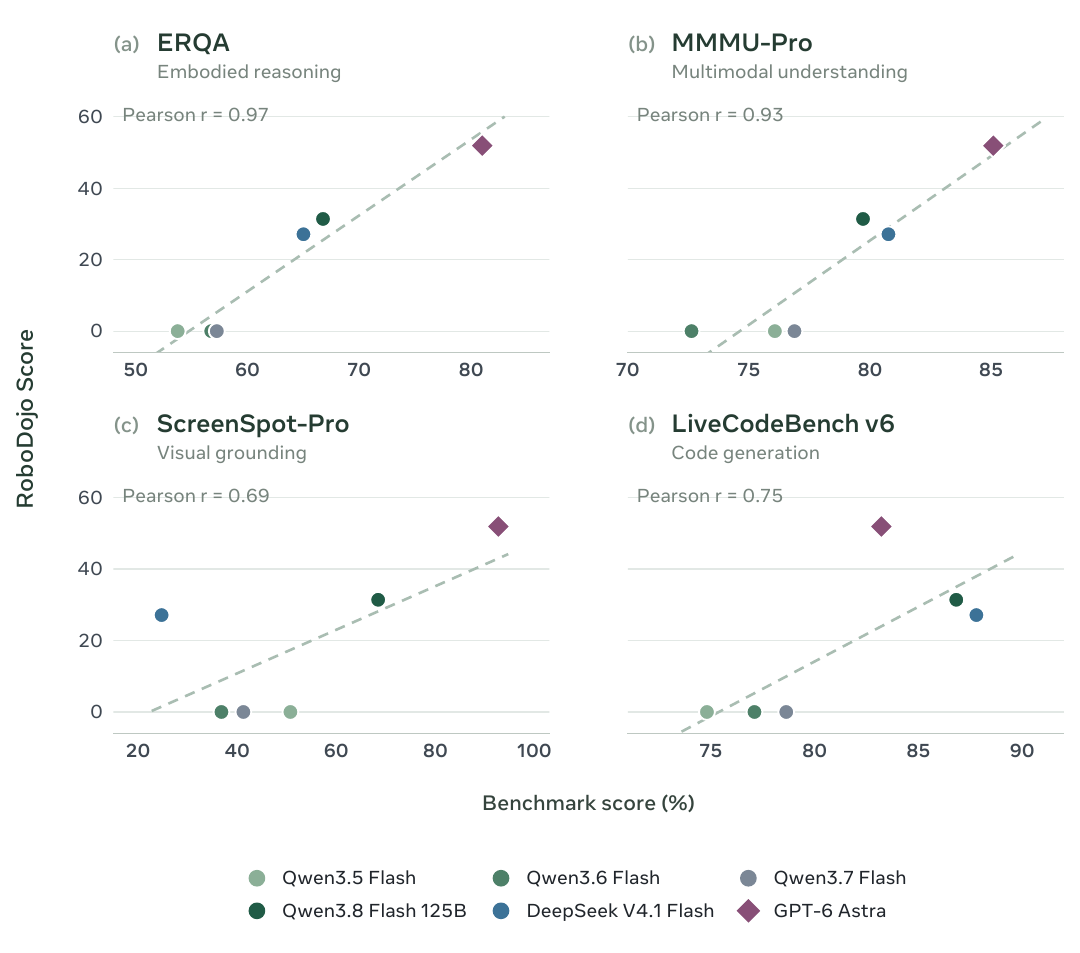}
  \caption{\textbf{Robot-use performance and general model capabilities.} Each point denotes a model, with benchmark performance on the horizontal axis and RoboDojo Score on the vertical axis. Model colors and symbols follow Figure~\ref{fig:four-failure-modes}. Dashed lines show linear fits; Pearson correlations are computed across the six models.}
  \label{fig:dojo-benchmark-correlations}
\end{figure}

\paragraph{Coding in geometry}
Coding skill is also closely correlated with task performance, as illustrated by geometry computations involving coordinate transformations and orientations. Earlier Qwen models frequently reverse coordinate transforms or mix quaternion conventions, with such errors affecting \textbf{47.3--81.8\%} of executed calculations across the earlier Flash and Plus cohorts. In contrast, the error rates are \textbf{8.2\%} for Qwen3.8-Flash, \textbf{1.5\%} for DeepSeek, and \textbf{0.7\%} for GPT-5.6 Sol, with no errors in GPT-6.1 Sol's calculations (Figure~\ref{fig:four-failure-modes}c).
% The pattern suggests that reliable robot-specific computation is another capability associated with effective robot use (Figure~\ref{fig:four-failure-modes}c).

\paragraph{Embodied reasoning.}
Beyond perceiving and computing a target pose, robot use requires embodied reasoning to propose appropriate interactions, including grasp poses. Because this broad capability is difficult to measure, we restrict our analysis to pick-and-place and use GPT-6 Astra to assess whether proposed grasp poses are physically reasonable. We report the rate at which the model proposed unnatural or infeasible grasping poses and failed to correct them in these tasks (Figure~\ref{fig:four-failure-modes}d). To our surprise, even recent models can lack this capability: GLM repeatedly fails to reorient its wrist for grasping, contributing to many unsuccessful trials and a \textbf{26.2\%} error rate. Such errors are less frequent in newer Qwen and GPT models, which more often select appropriate grasp orientations. However, the grasp-pose proposal is only one instance of embodied reasoning. GPT-6 Astra still struggles with other skills, including bimanual coordination, where inappropriate interaction plans limit performance. Embodied reasoning therefore appears to improve across model generations, but remains a substantial bottleneck beyond this narrow grasping analysis.

In conclusion, these failure patterns suggest that effective robot use draws on a combination of visual understanding, spatial grounding, geometric coding, and embodied reasoning, rather than a single robot-specific capability. This suggests that general model capabilities provide a substantial foundation for robot use. We therefore argue that future RUA training should develop these capabilities jointly -- for example, by using robot data to construct training tasks targeting each capability -- rather than focusing solely on fitting action outputs.

\subsection{Relationship to General Model Capabilities}
\label{sec:general-model-capabilities}

Having identified four key capabilities underlying effective robot use, we next ask: To what extent can these capabilities be captured by existing foundation-model benchmarks?

To investigate this, we compare RoboDojo performance with scores on four corresponding benchmarks: MMMU-Pro~\citep{yue2024mmmupro} for general visual understanding, ERQA~\citep{gemini2025robotics} for embodied reasoning, ScreenSpot-Pro~\citep{li2025screenspotpro} for spatial grounding, and LiveCodeBench v6~\citep{jain2024livecodebench} for code generation (Figure~\ref{fig:dojo-benchmark-correlations}). We observe positive correlations across all four benchmarks, with particularly strong correlations for embodied reasoning and general visual understanding, while spatial grounding exhibits a weaker correlation, potentially because ScreenSpot-Pro evaluates 2D screen-based grounding rather than the 3D spatial grounding required for robotic manipulation. These findings further support the role of general-purpose model capabilities in robot use, suggesting that advances in these capabilities may translate into improved robotic performance. %In addition, the common upward trend suggests that
% In practice, effective robot use usually draws on a combination of these capabilities: interpreting observations, locating relevant targets and reasoning about action consequences, and expressing decisions as executable code. These correlations further call for improving the four capabilities we identified together to enable more performant RUAs.

\vspace{-1.0em}
\section{Analysis on Privileged Information}
\label{sec:privileged-information}

%The preceding evaluation establishes the performance of \method across different models. 
As discussed in Section~\ref{sec:realistic-inputs}, privileged inputs such as precise simulator-provided depth and oracle task-success feedback can assist robot interaction. Although \method operates without these inputs, it remains important to understand how much they affect robot-use performance. We first challenge whether depth, a commonly used source of geometric information, is necessary for better performance (RQ1). Second, since skill libraries can encode privileged information or task-specific solutions, we ablate our skill library to test whether performance depends on such a leak of privileged information (RQ2).

\subsection{RQ1: How Does Depth Affect Robot Use?}
\label{sec:spatial-understanding}

%Identifying an object in an image does not directly tell an agent where to move the robot. 
We compare three levels of geometric information for depth: RGB images with depth and camera parameters, RGB images with camera parameters, and only RGB images. With depth and camera parameters, the agent can infer 3D coordinates directly from the pixel. Having only camera parameters requires the agent to find the pixel of corresponding points in multiple cameras. Without either, the agents should infer the coordinates through spatial relationships of the objects in the image.
We evaluate GPT-6 Astra and Qwen3.8-Flash-Next on the same 42 benchmark tasks.
The evaluation uses three seeds and two layouts per task for Qwen and one seed and one layout per task for Astra due to limited evaluation costs.

\begin{figure}[!t]
\centering
\includegraphics[width=0.98\linewidth]{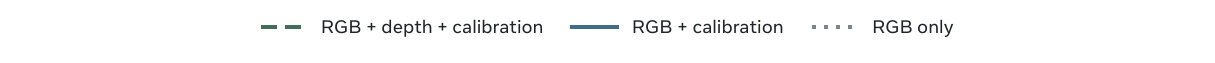}\par\smallskip
\captionsetup{subrefformat=parens}
\captionsetup[subfigure]{position=top,labelformat=parens,labelsep=space,font=footnotesize,labelfont=bf,textfont=bf,skip=4pt,justification=raggedright,singlelinecheck=false}
\begin{subfigure}[t]{0.32\linewidth}
\phantomsection\phantomcaption
\label{fig:geometry-astra-progress}
\includegraphics[width=\linewidth]{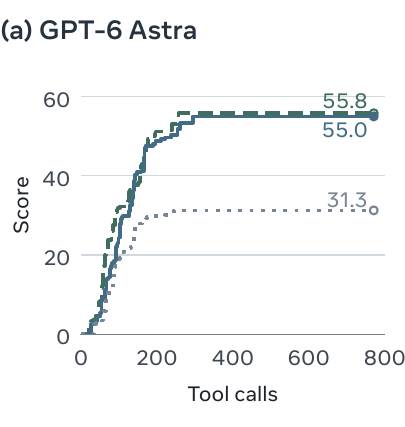}
\end{subfigure}\hfill
\begin{subfigure}[t]{0.32\linewidth}
\phantomsection\phantomcaption
\label{fig:geometry-qwen-progress}
\includegraphics[width=\linewidth]{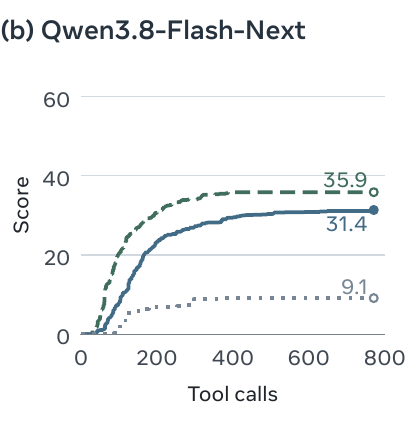}
\end{subfigure}\hfill
\begin{subfigure}[t]{0.32\linewidth}
\phantomsection\phantomcaption
\label{fig:geometry-information-trend}
\includegraphics[width=\linewidth]{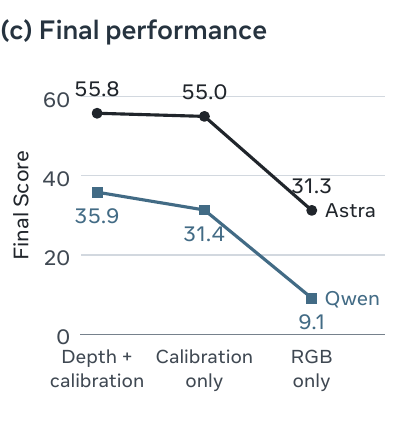}
\end{subfigure}
\caption{\textbf{Robot use with less geometric information.} Recorded progress through 771 tool calls and final score across 42 tasks, with skills enabled.}
\label{fig:depth-ablation}
  \par\vspace{\floatsep}\vspace{-4mm}
  \includegraphics[width=\linewidth]{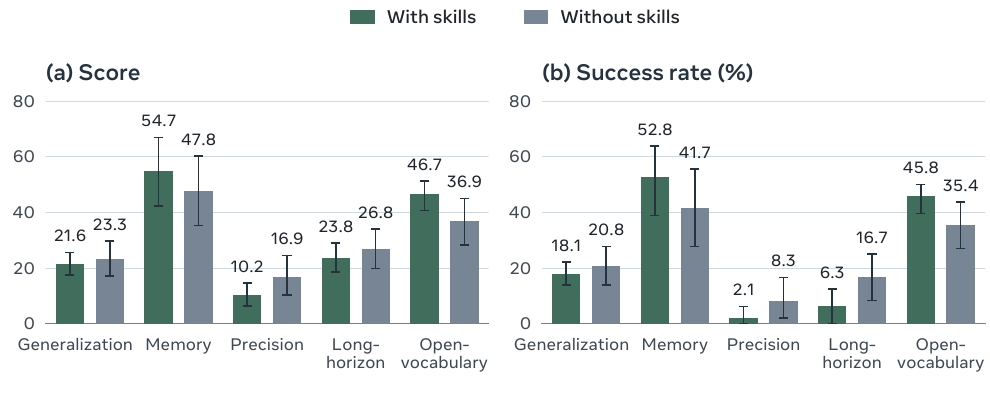}
  \caption{\textbf{Effect of skill guidance on task performance.} Qwen3.8-Flash-Next with \method, evaluated with and without skills. Error bars indicate approximate 95\% bootstrap confidence intervals.}
  \label{fig:skill-comparison}
  \vspace{-4mm}
\end{figure}

Without depth, agents can still recover object positions with the camera parameters by triangulating the corresponding points across views. Evidently, Astra's score changes from \textbf{55.83} to \textbf{55.04}, and Qwen's score changes from \textbf{35.89} to \textbf{31.39}, retaining 98.6\% and 87.5\% of their depth-enabled scores, respectively (Figures~\ref{fig:geometry-astra-progress} and~\ref{fig:geometry-qwen-progress}), reflecting a minimal loss in performance, especially with stronger model capabilities. 
The cases in Appendix~\ref{app:grounding-cases} further show that Qwen writes and executes its own geometry code to infer 3D coordinates, such as triangulating object positions across views and testing candidate table heights with the images.

Without camera parameters (only RGB images), where agents must infer 3D positions from abstract spatial relationships, Qwen's performance shows a sharp drop to \textbf{9.13}, while Astra shows a moderate decrease on its performance to \textbf{31.29} (Figure~\ref{fig:geometry-information-trend}).

In conclusion, we find that privileged depth information is \textbf{not necessary} for achieving strong robot-use performance as long as camera parameters are available. In fact, as model capabilities improve, models can more effectively infer 3D spatial relationships from visual observations, thereby benefiting less from such privileged inputs.

% Keep Figures 10 and 11 together at the top of page 17.

% Keep Figures 12 and 13 together at the top of page 18.
\begin{figure}[!t]
  \centering
  \includegraphics[width=\linewidth]{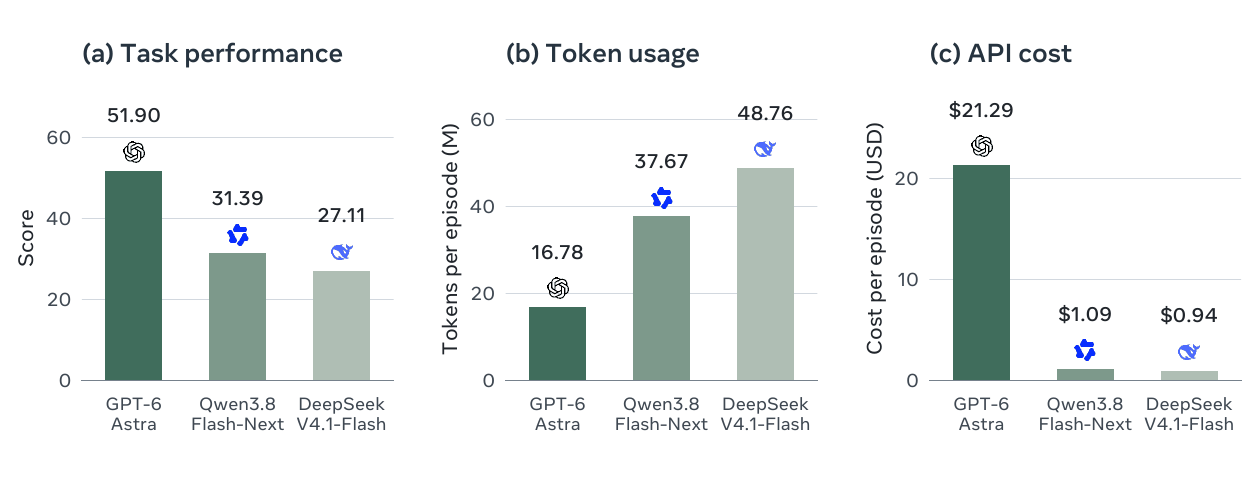}
  \caption{\textbf{Model-level resource--performance trade-offs.} Astra, Qwen, and DeepSeek each use 252 episodes. Bars show score and mean resource usage per episode.}
  \label{fig:model-resource-performance}
  \par\vspace{\floatsep}\vspace{-4mm}
  \includegraphics[width=\linewidth]{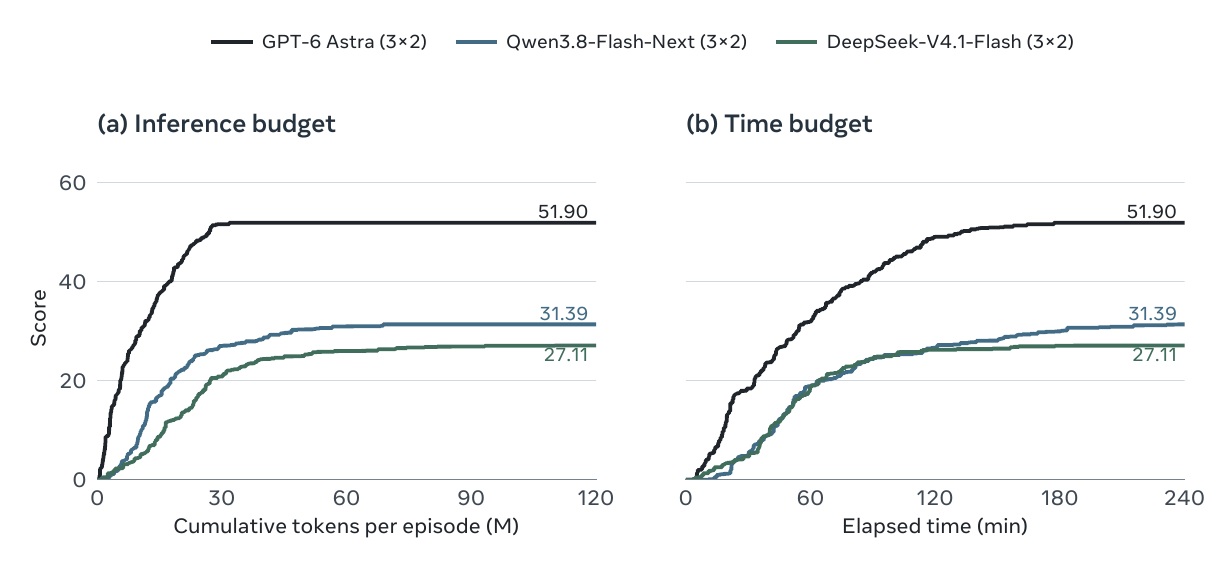}
  \caption{\textbf{Task progress over token usage and time.} The score is aggregated from recorded intermediate outcomes for the same model cohorts as Figure~\ref{fig:model-resource-performance}. Curves retain each episode's latest observed score.}
  \label{fig:model-budget-curves}
  \vspace{-3mm}
\end{figure}

\subsection{RQ2: Is skill library important for \method?}
\label{sec:task-generalization}

Task-specific skill libraries may encode privileged information or predefined solutions that simplify robot tasks. We therefore examine whether the performance of \method depends on the knowledge provided by its skill library.

Using Qwen3.8-Flash-Next, we disable access to the robot skill library and skill writing (we retain task instructions, environment descriptions, and standard robot tools). Figure~\ref{fig:skill-comparison} shows a small aggregate performance gap between configurations with and without robot skills. The no-skill configuration reaches \textbf{30.31} score and \textbf{24.58\%} SR, retaining \textbf{96.6\%} of the skill-enabled score (31.39), with an SR gap of only 0.42 percentage points.
The results show that \method maintains comparable performance without the skill library.
The harness supports reusable skills, but its performance does not depend on the task-specific knowledge.

\section{Analysis on Cost and Speed}
\label{sec:cost-speed}

In this section, we focus on the time and resources required for \method across different models, and show that pipelining of \method substantially improves the execution time.
\subsection{Token Usage and API Cost}
\label{sec:performance-cost}

A robot-use agent may spend many model calls and reasoning for locating objects, testing a grasp, or recovering from an unsuccessful action. We analyze the results from Section~\ref{sec:simulation-results}, examining how task performance relates to token usage, API cost, and progress during execution across different models. Reference API prices are provided in Appendix~\ref{app:cost-estimation}.

Figure~\ref{fig:model-resource-performance} shows that Qwen offers a favorable trade-off between the performance and the cost, reaching \textbf{31.39} score at \$1.09 per episode. Astra achieves the highest score (\textbf{51.90}), but costs nearly 20x more (\$21.29 per episode). This substantial difference of each model's cost highlights the value of advancing open-weight models for affordable robot-use agents.

Figure~\ref{fig:model-budget-curves} shows that each models need substantially different amount of tokens to make progress. To reach a score of 25, Astra requires 7.4M tokens per episode, which is less than 23.5M for Qwen and 49.9M for DeepSeek. This trend also persists in terms of the wall-clock time, where Astra reaches a score of 25 after 42.6 minutes, while Qwen and DeepSeek take 95.5 and 99.2 minutes, respectively.

\subsection{Execution Efficiency}
\label{sec:pipelining}

\begin{figure}[t]
  \centering
  \includegraphics[width=0.9\linewidth]{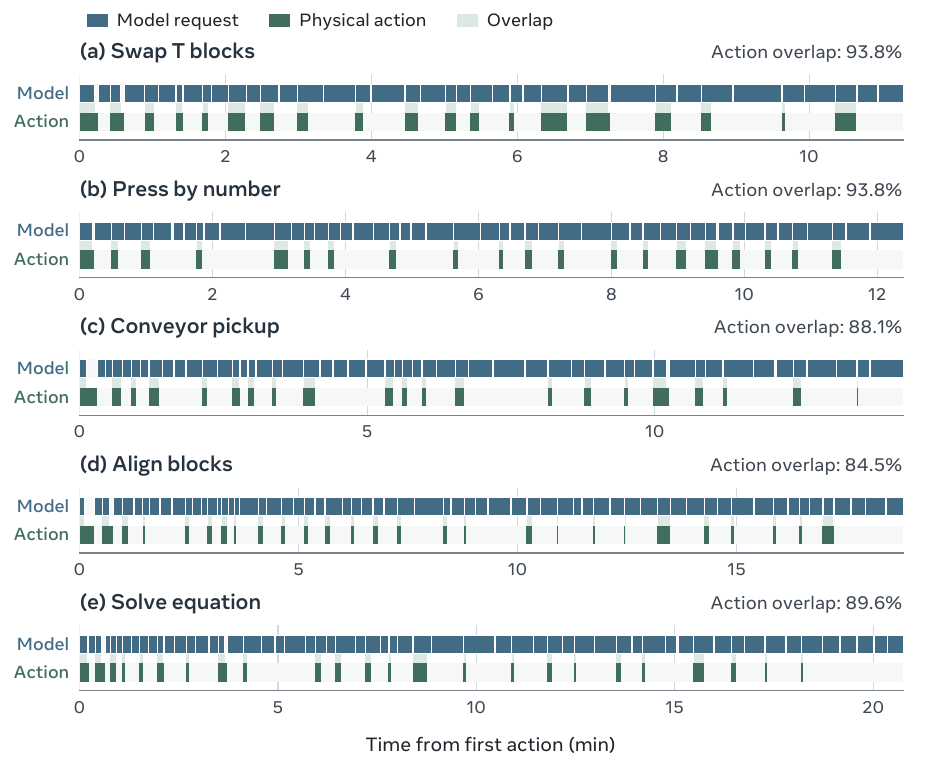}
  \caption{\textbf{Model--action overlap in pipelined execution.} Blue and green bars show model requests and physical actions; shading marks overlap. Percentages indicate the fraction of action time overlapping model requests. Each panel starts at the first action and uses its own time range.}
  \label{fig:execution-overlap}
\end{figure}

\begin{figure}[t]
  \centering
  \includegraphics[width=\linewidth]{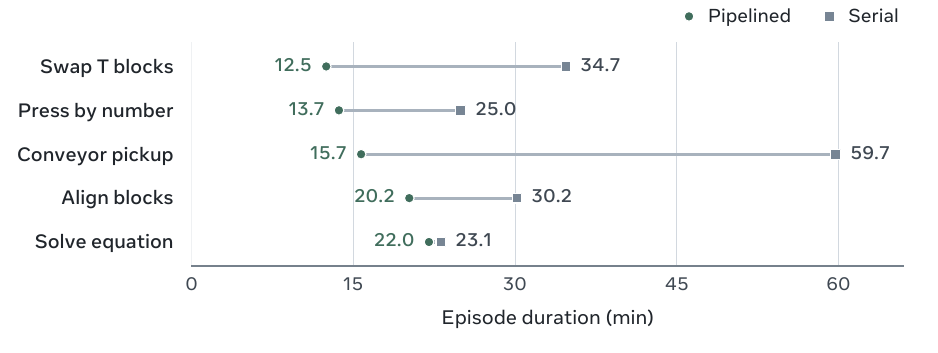}
  \caption{\textbf{Episode duration on five selected tasks.} Pipelined and serial execution are compared using one episode per configuration and task, excluding simulator initialization.}
  \label{fig:pipelining-comparison}
  \vspace{-4mm}
\end{figure}

The pipeline introduced in Section~\ref{sec:efficient-execution} allows model interaction to continue while physical actions is being executed. 
We select five tasks and compare pipelined and serial execution of \method using GPT-6 Astra with medium reasoning effort. Each task is evaluated once per configuration with the same seed and layout. We measure episode duration excluding simulator initialization and the overlap between model requests and physical tool execution.

\paragraph{Execution Overlap.}
We measure the fraction of physical execution time that overlaps with a model request. As shown in Figure~\ref{fig:execution-overlap}, it ranges from \textbf{84.5\% to 93.8\%} in the pipelined episodes and is zero in all five serial episodes, showing that most of the physical execution time is pipelined with the model inference. 
Appendix~\ref{app:pipelining-cases} shows additional qualitative case studies.

\paragraph{Episode Duration.}
As shown in Figure~\ref{fig:pipelining-comparison}, the pipelined runs have shorter episode durations on all five tasks, with observed reductions ranging from \textbf{4.8\% to 73.7\%} relative to serial execution. For example, \texttt{swap\_T} takes 12.5 minutes with pipelining versus 34.7 minutes with serial execution, while \texttt{press\_by\_number} takes 13.7 versus 25.0 minutes.

\section{Related Work}
\label{sec:related-work}
\paragraph{Vision-Language-Action and World-Action Models.}
Vision-language-action (VLA) models extend robot policies beyond task-specific demonstrations by transferring pretrained visual-language representations and learning from diverse robot data~\citep{brohan2023rt2,kim2024openvla}. Training on heterogeneous sources has further improved generalization: $\pi_{0.5}$ performs manipulation in previously unseen environments by combining robot experience with semantic supervision and web data~\citep{pi2025pi05}. World-action models (WAMs) explore a complementary route through predictive modeling. Joint video and action generation enables DreamZero to generalize to unseen tasks and environments~\citep{ye2026dreamzero}, while Cosmos Policy uses predicted future states and values to support test-time planning~\citep{kim2026cosmospolicy}. These advances broaden where learned policies can operate and what behaviors they can perform. Reliable use of past observations and recovery from failures over extended interactions nevertheless remain important areas for further investigation.

\paragraph{Tool-Using Agents for Robot Control.}
Tool use allows language models to organize decisions around observations and feedback. In computer environments, agents operate graphical interfaces through repeated observation and action, combining visual grounding with planning and context management~\citep{anthropic2024computeruse,wang2025uitars2,agashe2025agents2}. Robotics research has explored related ideas through skill selection and programmatic composition of perception and control APIs~\citep{ahn2022saycan,liang2022code}. Environment feedback enables replanning~\citep{huang2022innermonologue}, while spatial representations connect language instructions to physical motion~\citep{huang2023voxposer}. Together, these studies establish ways for general-purpose models to direct interaction through tools. In robotics, this interaction must also account for a world that changes while actions are running, making the coordination of observation, reasoning, and execution a central concern.

\paragraph{Robot-Use Agents and Harnesses.}
Recent studies show that general-purpose multimodal models can operate robots through tools without task-specific robot training~\citep{isola2026robotuse,zhang2026unexpected}. This has prompted efforts to design the harness through which a model observes the scene, issues actions, and receives feedback. RoboDawn and Robo-Harness K1 provide motion and geometric perception tools~\citep{guo2026robodawn,li2026roboharness}, while OpenRUA gives coding agents access to native ROS~2 interfaces through a terminal~\citep{chu2026openrua}. Beyond these interfaces, Harness VLA and GAP investigate how agents can guide learned policies through memory-based planning and corrections to policy-generated actions~\citep{zhang2026harnessvla,su2026astra}. ORA also demonstrates robot use with open-weight models, combining robot tools with environment information and context management~\citep{yang2026ora}. As robot use expands to more models and environments, reusing these components becomes an important system-design question. We develop \method around this need, separating shared robot tools from model runtimes and environment adapters so that changing one does not require rebuilding the others. We also coordinate tool execution and feedback to let agents continue reasoning while physical actions are underway.

\section{Conclusion}
\label{sec:conclusion}
We presented \method, a unified and modular harness with efficient execution and reliable tools for robot-use agents. Evaluations demonstrate strong performance with GPT-6 Astra in both simulation and real-world manipulation. More importantly, our results show that an open-weight model can also achieve strong robot-use performance: Qwen3.8-Flash-Next reaches a RoboDojo Score of 31.39, exceeding the published Astra--RoboProbe result of 28.97. These findings support further investigation of open-weight models that can be deployed locally and adapted through training on robot experience. Precise manipulation and reliable long-horizon execution remain challenges in our evaluations. By providing shared interaction interfaces, replaceable components, and structured execution feedback, \method offers a unified foundation for collecting robot interaction data, developing training methods, and evaluating their effects on agent behavior and task performance.

\bibliographystyle{abbrvnat}
\bibliography{main}

\clearpage
\appendix

\section{Experimental Setup and Reproducibility Settings}
\label{app:reproducibility-settings}

\paragraph{Simulation benchmark and baselines.}
RoboDojo provides 42 tasks in Isaac Sim~\citep{chen2026robodojo}. We compare agent harnesses using Qwen3.8-Flash-Next as the planner. GPT-as-Policy (GAP)~\citep{su2026astra} uses its hybrid configuration with $\pi_{0.5}$; RoboProbe~\citep{zhang2026unexpected} and RPent~\citep{zhang2026harnessvla} use their default configurations. Table~\ref{tab:robodojo-main} also includes the published Astra--RoboProbe results and representative vision-language-action (VLA) and world-action models (WAM) from the RoboDojo leaderboard~\citep{robodojo2026leaderboard}.

\paragraph{Reasoning effort.}
We use \texttt{medium} reasoning effort for GPT-6 Astra, \texttt{high} for both GPT-5.6 Sol and GPT-6.1 Sol, \texttt{xhigh} for all Qwen-series Models, and \texttt{max} for DeepSeek-V4.1-Flash.

% Additional inference and implementation settings will be documented here.

% Source workbook title: RealGen 真机测评, revision 747, read on 2026-10-20.
% https://ocnlxrdhjocp.feishu.cn/wiki/Xr42wmgU1iDmzkkUjFJc8C2enfh?sheet=8fcabb
% Test sheet: 8fcabb, rows 1--25; training sheet: XDLH1N, rows 1--31.
% Preserve the source's 19 evaluation entries and stated score caps.
\section{Real Eval Benchmark Details}
\label{app:real-eval-benchmark}

Real Eval evaluates tabletop manipulation on a dual-arm gripper platform
through five capability dimensions: \emph{spatial understanding},
\emph{object generalization}, \emph{instruction following},
\emph{compositional generalization}, and \emph{unseen tasks}. Its test
catalogue contains 19 evaluation entries, with 20 points allocated to each
dimension and 100 points in total. An evaluation entry specifies an operation
together with the variation being tested: bowl stacking, for example, appears
both as an object-generalization test and as a direction-following test. The
catalogue therefore distinguishes changes in objects, language, and task
structure even when they share a manipulation primitive.
Table~\ref{tab:real-eval-tasks} gives the complete test catalogue and its
scoring criteria.

\subsection{Hardware and Baselines}

All methods run on an ARX AC one dual-arm platform with two X5 arms and three Intel RealSense D405 cameras, comprising a head view and a wrist view on each arm. We evaluate \method with GPT-6 Astra and Qwen3.8-Flash-Next as planners and compare both configurations with $\pi_{0.5}$~\citep{pi2025pi05}, fine-tuned on demonstrations from the same platform. We also include Dream-1~\citep{dreamlabs2026dream1}, a world action model evaluated zero-shot through its public API (\texttt{dreamlabs/dream-1-beta}). In the real-world evaluation, \method additionally uses the D405 depth stream, whereas $\pi_{0.5}$ and Dream-1 natively consume only RGB images.

\subsection{Training Task Catalogue and Collection Protocol}
The accompanying training catalogue lists 30 task templates and 1,550
trajectories: 27 templates specify 50 trajectories each, while bag zipping,
coin insertion, and tissue extraction specify 100, 80, and 20, respectively.
The templates span stacking and sorting, object transfer and handover,
container loading, cloth manipulation, surface wiping, and operation of
buttons, switches, drawers, and a microwave door. Collection protocols vary
object placement, material, color, container arrangement, and table appearance
at task-specific intervals. For example, paper-cup stacking uses four to six
inverted cups at different positions, changes cup material every five
trajectories, and changes the table mat every ten. Test-tube transfer uses three
to four tubes of different sizes and reverses the transfer direction every
five trajectories. Cloth folding includes two folds per trajectory, changes
cloth orientation every three trajectories, and changes the cloth every six.
These collection settings establish the reference operations and variations
against which the test catalogue defines new objects, instructions, and action
combinations. The catalogue is collected to fine-tune end-to-end embodied
policies such as $\pi_{0.5}$, whereas \method does not use any of these
demonstrations; for \method, every entry is therefore evaluated zero-shot, and
the five dimensions characterize the type of variation that each entry
introduces.

\subsection{Evaluation Dimensions}

\paragraph{Spatial understanding.}
This dimension includes one simple entry (fruit transfer), two intermediate
entries (cloth folding and plate extraction), and two complex entries
(test-tube transfer and mug hanging). Fruit transfer moves cherries, bananas,
apples, oranges, and melons from bowls or plates into a basket. Cloth folding
varies size and orientation. Plate extraction requires one arm to lift a
bottle while the other removes and stacks the plate beneath it. Test-tube
transfer and mug hanging further distinguish acquiring an object from
completing the subsequent placement.

\paragraph{Object generalization.}
The operation remains bottle handover or bowl stacking while the manipulated
objects change. Handover introduces tea, coconut-water, and bottled-water
containers, together with handled stainless-steel and plastic cups. Bowl
stacking introduces stainless-steel bowls, small and large paper bowls,
ceramic bowls, and large salad bowls. These variations probe differences in
shape, size, material, and available grasp geometry.

\paragraph{Instruction following.}
Three tests separate color, object, and spatial-language grounding.
Color-conditioned cup stacking places two or three differently colored cups
on the table and specifies which color should be stacked onto which other
color. Object-conditioned pick-and-place uses a toy car, five fruit types,
blocks, and three tableware types, with bowls, plates, and baskets as target
containers. Direction-conditioned bowl stacking requests left-to-right or
right-to-left arrangements. These tests evaluate whether a physically
feasible operation is directed toward the requested object, destination, and
spatial relation.

\paragraph{Compositional generalization.}
Four entries recombine familiar operations, objects, and semantic constraints.
Tissue extraction followed by disposal chains two actions into one objective.
Stacking upward-facing cups changes the object configuration relative to the
inverted-cup training template. Wiping a whiteboard with a rag combines the
wiping operation with an object used elsewhere in the task catalogue, replacing
the whiteboard eraser specified for training. Arranging cubes by color changes
the criterion from the size ordering specified in the training catalogue.
Together, these cases test action sequencing, object substitution, and
semantic recombination.

\paragraph{Unseen tasks.}
The final dimension organizes transfer by its relation to the training
catalogue. Closing the microwave door isolates a sub-action of the microwave
loading routine. Taking a bowl out of the microwave and unfolding a rag
reverse the corresponding loading and folding operations. Stacking books is
listed as a combined action, while building a three-level Tower of Hanoi is
listed as a new action.

Representative tasks across the five dimensions are illustrated in
Figure~\ref{fig:real-world-tasks}.

\begingroup
\small
\renewcommand{\arraystretch}{1.08}
% Local dashed rules avoid arydshln's conflict with colored tabularx tables.
\def\RealEvalGroupRule{%
  \multispan{4}\xleaders\hbox{\rule{1.5pt}{0.4pt}\hspace{2pt}}\hfill\cr
  \noalign{\nobreak\vskip 2pt}}
\begin{longtable}{@{}c>{\raggedright\arraybackslash}p{0.25\linewidth}
                     >{\raggedright\arraybackslash}p{0.58\linewidth}r@{}}
\caption{Complete Real Eval test catalogue. Each dimension has a 20-point
budget, and score caps follow the source specification. IDs match
Table~\ref{tab:real-eval-per-task}.}
\label{tab:real-eval-tasks}\\
\toprule
ID & Task & Task variation and scoring criterion & Max \\
\midrule
\endfirsthead
\caption[]{Real Eval test catalogue (continued).}\\
\toprule
ID & Task & Task variation and scoring criterion & Max \\
\midrule
\endhead
\midrule
\multicolumn{4}{r@{}}{Continued on the next page.}\\
\endfoot
\bottomrule
\endlastfoot

\multicolumn{3}{@{}l}{\textbf{Spatial Understanding}} & \textbf{20}\\*[2pt]
\RealEvalGroupRule
1 & Pick and place (fruits) &
Move cherries, bananas, apples, oranges, and melons from bowls or plates
into a basket. One point for each fruit type placed successfully in every
trial. & 5\\
2 & Fold rag &
Fold large, small, and obliquely placed rags. One point for each condition
folded successfully. & 3\\
3 & Bimanual plate pull-out &
Lift the bottle with one arm and remove and stack the plate with the other.
Each arm's required action receives 0.5 points per trial. & 4\\
4 & Insert test tubes into rack &
Transfer test tubes from one rack and insert them into another. Each pickup
receives 0.5 points and each insertion 0.5 points. & 4\\
5 & Hang mug &
Hang a mug on a rack. Each grasp receives 0.5 points and each successful
hang 0.5 points. & 4\\
\addlinespace

\multicolumn{3}{@{}l}{\textbf{Object Generalization}} & \textbf{20}\\*[2pt]
\RealEvalGroupRule
6 & Bimanual bottle handover &
New beverage bottles and handled steel/plastic cups. One point per
successful handover, capped at ten. & 10\\
7 & Stack bowls (new objects) &
New steel, small/large paper, ceramic, and large salad bowls. One point per
successful stack, capped at ten. & 10\\
\addlinespace

\multicolumn{3}{@{}l}{\textbf{Instruction Following}} & \textbf{20}\\*[2pt]
\RealEvalGroupRule
8 & Stack cups (color order) &
Two- and three-color settings; one point per correctly distinguished color
configuration, capped at five points for each setting. & 10\\
9 & Pick and place (assorted objects) &
Select among a toy car, five fruits, blocks, and three tableware types;
place into a bowl, plate, or basket. Each distinguished object and each
container category receives 0.5 points. & 6.5\\
10 & Stack bowls (given direction) &
Follow left/right and above/below relations. Left/right discrimination
receives two points; above/below receives 1.5. & 3.5\\
\addlinespace

\multicolumn{3}{@{}l}{\textbf{Compositional Generalization}} & \textbf{20}\\*[2pt]
\RealEvalGroupRule
11 & Pull out tissue and discard &
Pull a tissue out of its pack and throw it into a bin. Each extraction and
each disposal receives 0.5 points, up to two points each. & 4\\
12 & Stack upward-facing cups &
Stack paper cups with their openings facing upward. One point per
successful stack. & 4\\
13 & Wipe whiteboard &
Use a rag to wipe the whiteboard. One point per successful wipe. & 4\\
14 & Arrange cubes (color order) &
Arrange cubes according to color in two- and three-color settings; one point
per correctly distinguished configuration, capped at four points per
setting. & 8\\
\addlinespace

\multicolumn{3}{@{}l}{\textbf{Unseen Tasks}} & \textbf{20}\\*[2pt]
\RealEvalGroupRule
15 & Close microwave door &
Sub-action transfer: close the microwave door. One point per success. & 4\\
16 & Take bowl from microwave &
Inverse action: take a bowl out of the microwave. One point per
success. & 4\\
17 & Unfold rag &
Inverse action: unfold the rag. One point per success. & 4\\
18 & Stack books &
Combined action: stack notebooks. One point per success. & 4\\
19 & Tower of Hanoi &
New action: build a three-level Tower of Hanoi. One point per
success. & 4\\
\end{longtable}
\endgroup
\subsection{Per-Task Results}
\label{app:real-eval-results}
Table~\ref{tab:real-eval-per-task} reports the individual task scores underlying
Table~\ref{tab:real-eval-summary}, using the task IDs and scoring criteria above.

\subsection{Scoring Protocol}
\label{app:real-eval-scoring}
Before each trial, the evaluator resets the scene and randomizes object
placements and goal regions. Each task provides a pool of paraphrased
instructions; the evaluator selects one per trial and passes it to the method
as the only task specification. All methods, including $\pi_{0.5}$ and
\method, receive the same task-level instruction, without task-specific
demonstrations, subtask plans, or hints. After each trial, the evaluator
records whether each scoring criterion is satisfied, and all methods are
scored with the same rubric.

Let $\mathcal{T}_d$ denote the evaluation entries in dimension $d$, and let
$p_t$ and $m_t$ be the awarded points and stated maximum for entry $t$.
The benchmark's score allocation is
\begin{equation}
  S_d = \sum_{t\in\mathcal{T}_d} p_t, \qquad
  \sum_{t\in\mathcal{T}_d} m_t = 20, \qquad
  \mathrm{Score}_{\mathrm{Real\,Eval}} = \sum_{d=1}^{5} S_d,
  \label{eq:real-eval-score}
\end{equation}
with $0\leq p_t\leq m_t$ and a total score on a 0--100 scale. Equal dimension
budgets keep all five capabilities represented, while the individual task
weights follow the source rubric. Instruction tests additionally award credit
for correctly distinguished colors, objects, containers, or directions. Thus,
the aggregate is a rubric-based capability score, rather than the fraction of
episodes that fully complete their objectives. In particular, successful
grasping followed by failed insertion can receive partial credit without
constituting complete task success.

\begin{table*}[!t]
\centering
\caption{\textbf{Per-task results on Real Eval.} Each capability subtotal is out of 20, and Total is out of 100. Task IDs correspond to Table~\ref{tab:real-eval-tasks}; scoring details are given in Appendix~\ref{app:real-eval-scoring}.}
\label{tab:real-eval-per-task}
\fontsize{8}{10}\selectfont
\setlength{\tabcolsep}{4pt}
\renewcommand{\arraystretch}{1.2}
\begin{tabularx}{\textwidth}{@{}c l c
  *{2}{>{\centering\arraybackslash\hsize=0.8\hsize}X}
  *{2}{>{\centering\arraybackslash\hsize=1.2\hsize}X}@{}}
\toprule
 & & & \textbf{VLA} & \textbf{WAM} & \multicolumn{2}{c}{\textcolor{paperAccent}{\textbf{\method}}} \\
\cmidrule(lr){4-4}\cmidrule(lr){5-5}\cmidrule(l){6-7}
\textbf{ID} & \textbf{Task} & \textbf{Max}
 & $\pi_{0.5}$ & Dream-1
 & \multicolumn{1}{c}{GPT-6 Astra} & \multicolumn{1}{c}{Qwen3.8-Flash-Next} \\
\midrule
\multicolumn{7}{@{}l}{\textbf{Spatial Understanding}} \\
1  & Pick and place (fruits)           & 5   & 2    & 3   & 5    & 3   \\
2  & Fold rag                          & 3   & 1.75 & 1.5 & 3    & 1   \\
3  & Bimanual plate pull-out           & 4   & 3    & 0   & 2    & 1   \\
4  & Insert test tubes into rack       & 4   & 2    & 1   & 3    & 2   \\
5  & Hang mug                          & 4   & 2.5  & 1   & 4    & 2   \\
\rowcolor{paperAccentSoft}
   & \textit{Subtotal}                 & \textit{20} & \textit{11.25} & \textit{6.50} & \textit{17.00} & \textit{9.00} \\
\midrule
\multicolumn{7}{@{}l}{\textbf{Object Generalization}} \\
6  & Bimanual bottle handover          & 10  & 2.5  & 5.5 & 10   & 4   \\
7  & Stack bowls (new objects)         & 10  & 10   & 8   & 7.5  & 6   \\
\rowcolor{paperAccentSoft}
   & \textit{Subtotal}                 & \textit{20} & \textit{12.50} & \textit{13.50} & \textit{17.50} & \textit{10.00} \\
\midrule
\multicolumn{7}{@{}l}{\textbf{Instruction Following}} \\
8  & Stack cups (color order)          & 10  & 6    & 1   & 8    & 5   \\
9  & Pick and place (assorted objects) & 6.5 & 6    & 6.5 & 6.5  & 6.5 \\
10 & Stack bowls (given direction)     & 3.5 & 2    & 2   & 3.5  & 2   \\
\rowcolor{paperAccentSoft}
   & \textit{Subtotal}                 & \textit{20} & \textit{14.00} & \textit{9.50} & \textit{18.00} & \textit{13.50} \\
\midrule
\multicolumn{7}{@{}l}{\textbf{Compositional Generalization}} \\
11 & Pull out tissue and discard       & 4   & 0    & 4   & 4    & 1   \\
12 & Stack upward-facing cups          & 4   & 4    & 0   & 4    & 0   \\
13 & Wipe whiteboard                   & 4   & 2    & 4   & 4    & 4   \\
14 & Arrange cubes (color order)       & 8   & 0    & 2   & 8    & 8   \\
\rowcolor{paperAccentSoft}
   & \textit{Subtotal}                 & \textit{20} & \textit{6.00} & \textit{10.00} & \textit{20.00} & \textit{13.00} \\
\midrule
\multicolumn{7}{@{}l}{\textbf{Unseen Tasks}} \\
15 & Close microwave door              & 4   & 4    & 4   & 4    & 4   \\
16 & Take bowl from microwave          & 4   & 0    & 0   & 0    & 0   \\
17 & Unfold rag                        & 4   & 0    & 0   & 0    & 0   \\
18 & Stack books                       & 4   & 0    & 0   & 2    & 4   \\
19 & Tower of Hanoi                    & 4   & 0    & 0   & 2    & 0   \\
\rowcolor{paperAccentSoft}
   & \textit{Subtotal}                 & \textit{20} & \textit{4.00} & \textit{4.00} & \textit{8.00} & \textit{8.00} \\
\midrule
   & \textbf{Total}$\uparrow$          & \textbf{100} & 47.75 & 43.50 & \textbf{80.50} & 53.50 \\
\bottomrule
\end{tabularx}
\vspace{5pt}
\end{table*}

\section{API Cost Estimation}
\label{app:cost-estimation}

Fewer tokens do not necessarily mean lower cost: models differ in their prices for input, cached input, and output. For the comparisons in Section~\ref{sec:performance-cost}, we estimate API cost by applying the public rates in Table~\ref{tab:api-pricing} to each request's recorded usage and summing the charges over an episode. Uncached and cached input are billed separately. Reasoning tokens are already included in output usage and are counted only once. Recorded retries are included; requests without recoverable usage are left unestimated.

\begin{table}[ht]
\centering
\caption{\textbf{Reference API rates for the cost analysis.} Rates are per million tokens in the listed currency.}
\label{tab:api-pricing}
\small
\setlength{\tabcolsep}{5pt}
\begin{tabular}{@{}llrrr@{}}
\toprule
Model / pricing condition & Currency & Input & Cached input & Output \\
\midrule
GPT-6 Astra~\citep{openai2026astrapricing} & USD & 10.00 & 1.00 & 50.00 \\
Qwen3.8-Flash~\citep{alibaba2026qwenpricing} & CNY & 0.80 & 0.10 & 2.70 \\
DeepSeek-V4.1-Flash (peak)~\citep{deepseek2026pricing} & CNY & 2.00 & 0.04 & 8.00 \\
DeepSeek-V4.1-Flash (off-peak) & CNY & 1.00 & 0.02 & 4.00 \\
\bottomrule
\end{tabular}
\par\vspace{4pt}
\begin{minipage}{\linewidth}
\footnotesize
Rates were archived on October 22--23, 2026. Qwen uses the Beijing-region \texttt{qwen3.8-flash} rates. All costs are reported in USD, with CNY prices converted at 6.6729 CNY per USD.
\end{minipage}
\end{table}

\noindent\textbf{Astra.} Long-context requests and cache writes require additional pricing rules. Requests exceeding 272K input tokens are charged at twice the input and cache rates and 1.5 times the output rate. For recorded cache writes, we use the 30-minute rate of USD 12.50 per million tokens, also applying this rate when the retention duration is unspecified.

\noindent\textbf{DeepSeek.} The applicable rate depends on when a request is processed. We use the recorded response time, falling back to the request start time if it is unavailable. Peak rates apply from 09:00--12:00 and 14:00--18:00 on working days in the Asia/Shanghai time zone; all other times use off-peak rates. None of the evaluated requests fall on public holidays.

\section{Monitored Motion Execution}
\label{app:motion-execution}

The execution monitoring described in Section~\ref{sec:key-design-choices} addresses a basic difficulty in robot control: issuing a motion command does not establish that the robot has reached its target. Planning may reject a command, or execution may stop making progress. The \texttt{move} tool therefore advances through bounded waypoints, checks the state returned after each step, and reports why execution ended so that the agent can decide what to do next.

Algorithm~\ref{alg:monitored-move} gives this loop for a single active arm without a gripper-opening target. We denote the requested pose by $g=(p_g,q_g)$ and the execution target by $\tilde g$, initially equal to $g$. The two are kept separate because a supported planning rejection may lead to an alternative execution target within the original tolerances. Other arms retain their initial pose and gripper commands. For multi-arm requests, each active arm has its own pose tracker and all target poses must be reached; optional gripper-opening checks are omitted from the pseudocode.

\begin{algorithm}[!htb]
\caption{Monitored execution of a pose-only \texttt{move} request}
\label{alg:monitored-move}
\begin{algorithmic}[1]
\small
\Require Target $g$, tolerances $(\epsilon_p,\epsilon_q)$, waypoint limit $\delta$, rotation radius $r$, accepted-step budget $B$
\Ensure Actual pose, target errors, arrival flag, step count, stop reason, and diagnostic evidence
\State $s\gets\Call{ReadState}{}$; $\tilde g\gets g$; $k\gets0$; $a\gets\mathrm{false}$
\State Initialize progress history $H$ from $s$ and $\tilde g$; $z\gets\texttt{max\_steps}$
\While{$k<B$}
    \State \textbf{if} $s$ has ended \textbf{then} $z\gets\texttt{native\_end}$; \textbf{break}
    \State $w\gets\Call{Waypoint}{s,\tilde g,\delta,r}$
    \State $(s',b,d)\gets\Call{Step}{w}$ \Comment{New state, acceptance flag, diagnostics}
    \State $u\gets\Call{AbnormalMotion}{s,s'}$; $s\gets s'$
    \State \textbf{if} $b$ \textbf{then} $k\gets k+1$
    \State \textbf{if} $s$ has ended \textbf{then} $z\gets\texttt{native\_end}$; \textbf{break}
    \State \textbf{if} $u$ \textbf{then} $z\gets\texttt{unsafe\_motion}$; retain motion evidence; \textbf{break}
    \If{\textbf{not} $b$}
        \If{\textbf{not} $a$ \textbf{and} \Call{EligibleGoalSearch}{$d$}}
            \State $a\gets\mathrm{true}$
            \State $g'\gets\Call{SolveGoalRegion}{s,g,\epsilon_p,\epsilon_q}$
            \If{a valid alternative goal $g'$ is found}
                \State $\tilde g\gets g'$; reset $H$ using $s$ and $\tilde g$
                \State \textbf{continue}
            \EndIf
        \EndIf
        \State $z\gets\texttt{planning\_rejected}$; retain rejection evidence; \textbf{break}
    \EndIf
    \State Update $H$ with position and orientation errors relative to $\tilde g$
    \If{\Call{Reached}{$s,g,\epsilon_p,\epsilon_q$}}
        \State $z\gets\texttt{pose\_reached}$; \textbf{break}
    \ElsIf{\Call{Diverging}{$H$}}
        \State $z\gets\texttt{diverging}$; retain progress evidence; \textbf{break}
    \ElsIf{\Call{Stalled}{$H,\epsilon_p,\epsilon_q$}}
        \State $z\gets\texttt{stalled}$; retain progress evidence; \textbf{break}
    \EndIf
\EndWhile
\State \Return \Call{Feedback}{$s,g,\tilde g,k,z$} with diagnostic evidence
\end{algorithmic}
\end{algorithm}

\paragraph{Waypoint generation.}
Each waypoint is computed from the latest measured pose, so the next command accounts for where the robot actually moved. Translation and rotation share a displacement budget $\delta$, with the arm's declared rotation radius $r$ used to account for displacement due to rotation. The angular step is capped at
\begin{equation}
\theta_{\max}=\begin{cases}
\min\{0.6,\,2\arcsin(\min\{1,\delta/(4r)\})\}, & r>0,\\
0.6, & r=0,
\end{cases}
\end{equation}
in radians. The orientation is interpolated toward $\tilde g$ within this limit. For the resulting rotation $\Delta\theta$, the remaining translation allowance is
\begin{equation}
\delta_p=\max\{0,\,\delta-2r\sin(\Delta\theta/2)\}.
\end{equation}
The position advances toward $\tilde g$ by at most $\delta_p$.

\paragraph{Stopping criteria.}
The tool checks both whether the requested pose has been reached and whether continued execution is making progress. Arrival requires position and orientation errors relative to $g$ to fall within $\epsilon_p$ and $\epsilon_q$. Position error is Euclidean distance; orientation error is the shortest angular distance between unit quaternions,
\begin{equation}
d_q(q_1,q_2)=2\arccos(\min\{1,|q_1^\top q_2|\}).
\end{equation}

Progress is measured toward the execution target $\tilde g$. Let $e^p_t$ and $e^q_t$ be the corresponding errors after accepted step $t$. To detect motion away from the target, \textsc{Diverging} tracks $v_t=e^p_t+0.1e^q_t$, with position in meters and orientation in radians, and stops after two consecutive increases satisfying $v_t>v_{t-1}+0.01$. To detect lack of progress, \textsc{Stalled} examines four accepted-step measurements and stops when neither error has improved sufficiently:
\begin{equation}
e^p_{t-3}-\min_{j=t-2}^{t}e^p_j\leq0.1\epsilon_p,
\qquad
e^q_{t-3}-\min_{j=t-2}^{t}e^q_j\leq0.1\epsilon_q.
\end{equation}

Large changes in measured pose are checked separately by \textsc{AbnormalMotion}. The thresholds are $0.2$\,m or $0.8$\,rad for an active arm, and $0.05$\,m or $0.25$\,rad for an inactive arm. These checks also apply to rejected commands: rejection does not guarantee that the returned state is unchanged. Execution also stops when the environment ends or the step budget $B$ is exhausted. Only accepted steps count toward $B$ or update the progress history.

\paragraph{Planning rejections and feedback.}
A planning rejection may allow the tool to try a nearby execution target without relaxing the agent's original request. On backends that support goal-region search, \textsc{EligibleGoalSearch} permits one search per \texttt{move} call following an active-arm rejection of type \texttt{ik\_failed}, \texttt{joint\_jump\_rejected}, or \texttt{ik\_candidate\_rejected}. An alternative goal is accepted only if it refers to the current state version and lies within the original position and orientation tolerances. It replaces $\tilde g$ and resets the progress history, while arrival continues to be checked against $g$.

The returned feedback tells the agent where the robot ended up, how far it remains from the requested pose, and why execution stopped. It includes actual and target poses, position and orientation errors, arrival status, accepted-step count, and available rejection or recovery diagnostics. Reaching a pose does not by itself establish that a grasp, placement, or other task objective has succeeded.

% Keep case-study figures with their discussion without separate float pages.
\begingroup
\setlength{\intextsep}{8pt plus 2pt minus 2pt}
\setlength{\textfloatsep}{12pt plus 2pt minus 2pt}
\captionsetup[figure]{skip=6pt}
\section{Spatial Grounding Through Agent-Written Code}
\label{app:grounding-cases}

\begin{figure}[!htb]
  \centering
  \includegraphics[width=.96\linewidth]{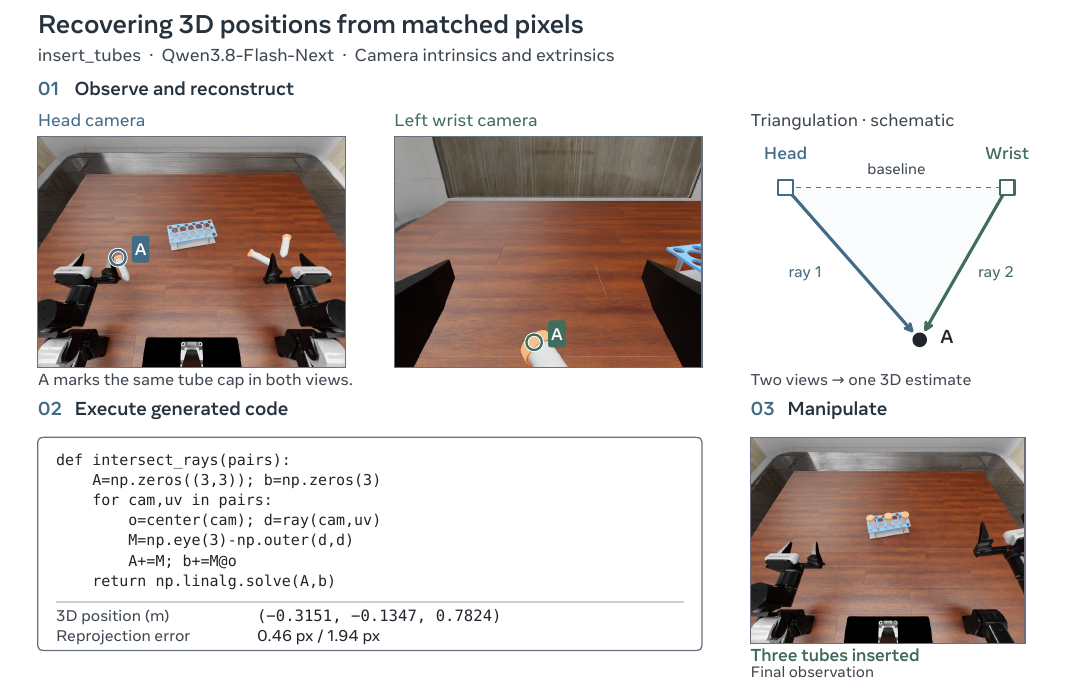}
  \caption{\textbf{Agent-written triangulation in \texttt{insert\_tubes}.} Matching pixels marked A define viewing rays, illustrated schematically above the executed code excerpt. Recorded coordinates and reprojection errors connect the calculation to its output; the final frame follows subsequent grasping and insertion.}
  \label{fig:depth-code-triangulation}
\end{figure}

The depth ablation in Section~\ref{sec:spatial-understanding} raises a practical question: how does an agent obtain usable spatial coordinates when its observations contain no depth? We examine two successful Qwen3.8-Flash-Next episodes from the no-depth, no-skill configuration. In both, the agent writes and executes geometry code using RGB observations and the supplied camera intrinsics and extrinsics, rather than calling the built-in triangulation operation. The cases illustrate different ways of constructing localization estimates: matching features across views and testing geometric assumptions against image evidence.

\subsection{Recovering Tube Positions Across Views}

In \texttt{insert\_tubes}, the agent must place three tubes into a rack. After obtaining image-space segmentation boxes, it captures additional camera views and writes a Python function to intersect viewing rays in a least-squares sense. It then selects corresponding tube-cap pixels and computes their world coordinates. Figure~\ref{fig:depth-code-triangulation} follows one cap: the calculated position reprojects within 0.46 and 1.94 pixels of the selected points in the two views. This gives the agent a way to check whether its chosen correspondences agree geometrically.

The cap positions are only the beginning of the manipulation sequence. The agent subsequently estimates tube axes, chooses grasp poses, and uses observations and action feedback to guide lifting and insertion. The episode ends with all three tubes in the rack. Here, code execution turns image correspondences into coordinates that the agent can inspect and use in its action planning.

\begin{figure}[!htb]
  \centering
  \includegraphics[width=.96\linewidth]{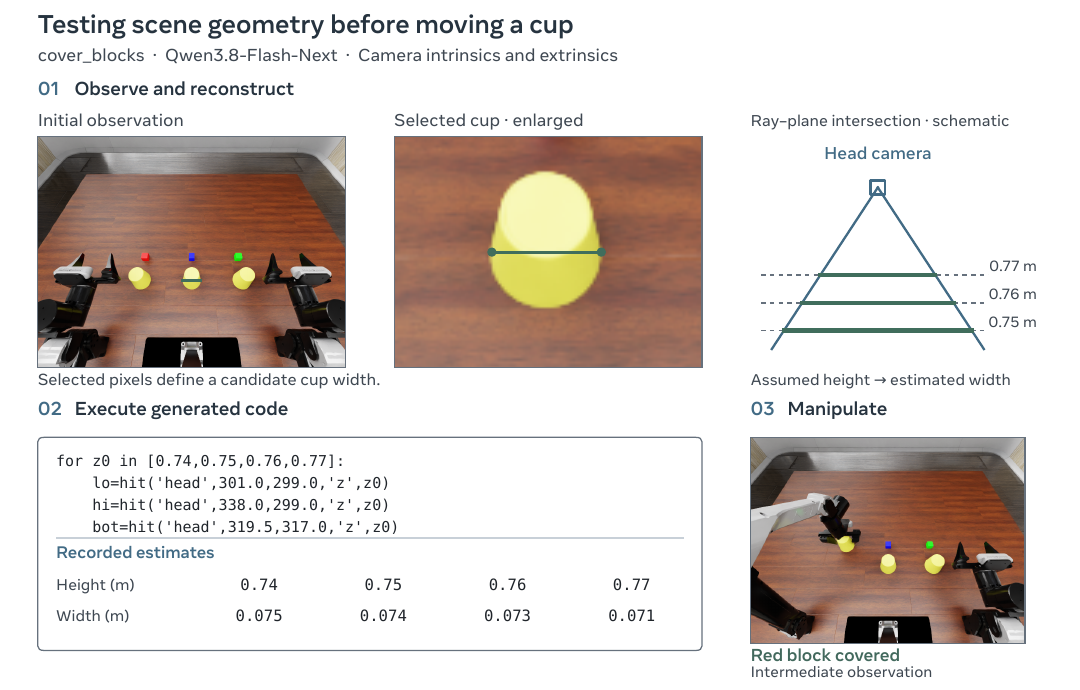}
  \caption{\textbf{Geometric hypothesis testing in \texttt{cover\_blocks}.} The agent varies a candidate table height and computes cup dimensions from selected pixels. The schematic illustrates the ray--plane intersections computed by \texttt{hit()} in the code excerpt; the right-hand frame shows a later placement over the red block.}
  \label{fig:depth-code-hypotheses}
  % \vspace{-4mm}
\end{figure}

\subsection{Exploring Geometry Before Moving a Cup}

In \texttt{cover\_blocks}, the agent must cover three colored blocks from left to right, then uncover them in a specified color order. Locating the cups requires reasoning about their extent as well as their image centers. The agent extracts colored image regions and writes code that tests candidate table heights, intersects pixel rays with each candidate plane, and compares the resulting cup dimensions (Figure~\ref{fig:depth-code-hypotheses}). The height sweep exposes how an assumed scene parameter changes the inferred geometry, rather than treating a single image estimate as fixed.

The agent continues by projecting candidate object shapes into the images and refining its scene estimates. It then uses the estimated cup positions to approach, grasp, and place the cups, checking the scene as the task proceeds. The recorded episode completes the required covering and uncovering sequence. This case shows a more exploratory use of code: the agent constructs and tests geometric hypotheses before committing to physical actions. Together, the two cases illustrate how RGB-based localization can support successful execution, while requiring correspondence selection, geometric assumptions, and verification that depth access can simplify.

\section{Planning During Physical Execution}
\label{app:pipelining-cases}

The overlap measurements in Section~\ref{sec:pipelining} show when model requests and physical actions run concurrently, but not what the agent prepares during that time or how it uses the result. We follow longer excerpts from two recorded trajectories to connect this preparation to subsequent actions and feedback. Each figure pairs key observations with the recorded execution timeline and enlarges one interval in which the model generates a concrete calculation while a physical tool is running. The annotations summarize recorded outputs; the bars denote model requests, not a measurement of internal reasoning time.

\begin{figure}[!htb]
  \centering
  \includegraphics[width=.94\linewidth]{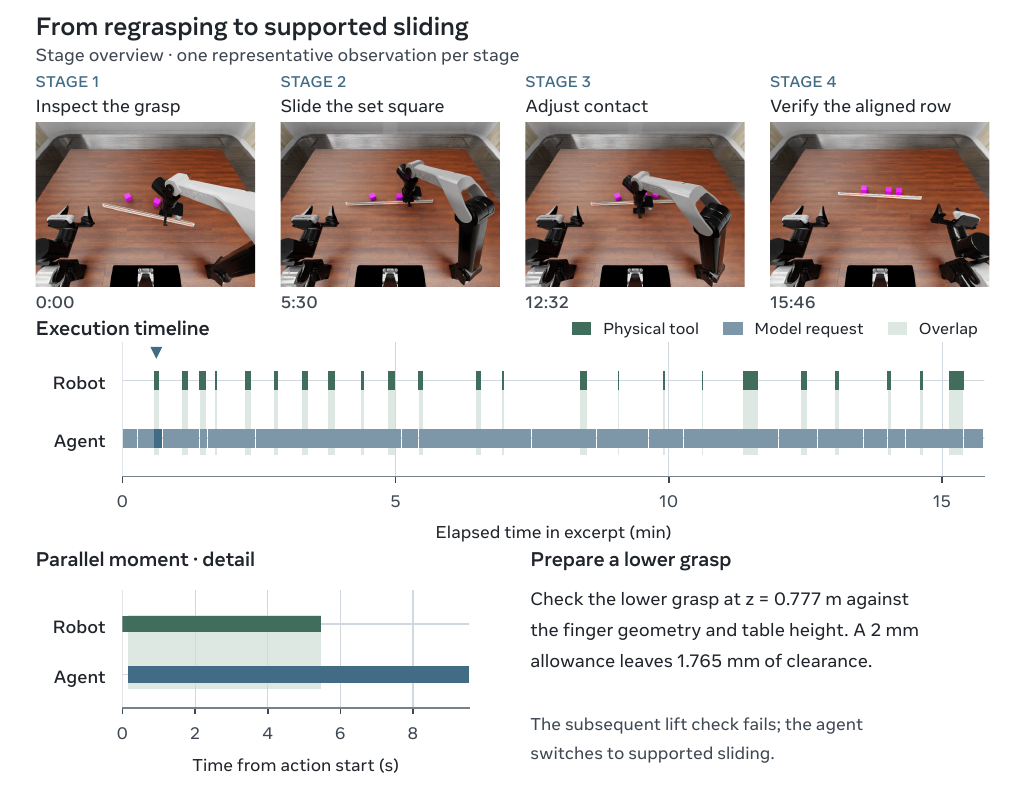}
  \caption{\textbf{Grasp adjustment and supported sliding in \texttt{align\_blocks}.} The upper row summarizes four stages, each illustrated by a representative observation; a stage may contain multiple tool calls. Green and blue bars show physical tools and model requests; shading marks overlap. The marked interval is enlarged below, where the model prepares a clearance check during release. Image timestamps and the main axis are relative to the first displayed observation.}
  \label{fig:pipelining-cases}
\end{figure}

\subsection{From Regrasping to Supported Sliding}
\label{app:case-align}

In \texttt{align\_blocks}, the agent must align three blocks with a set square while keeping the blocks on the table. Figure~\ref{fig:pipelining-cases} follows the trajectory from an uncertain grasp through contact adjustments to the final aligned row. While a background \texttt{release} opens the gripper, the model generates a Python check for a lower grasp pose. Using the finger geometry and table height, it checks that the proposed pose leaves clearance even after a 2\,mm uncertainty allowance. The program runs after the release finishes, and the agent then moves to the checked pose, grasps, and tests retention with a short lift.

The lift does not retain the set square. The agent explicitly changes its approach: it will keep the square supported and use sliding contact instead. It lowers the hand, closes the gripper, and issues a sequence of small rotations and translations, then inspects the scene and adjusts contact again. The final observation shows the blocks in a row with the arms clear. This longer sequence shows both the use of preparation during background execution and the need to revise that preparation when subsequent physical evidence contradicts it.

\begin{figure}[!htb]
  \centering
  \includegraphics[width=.94\linewidth]{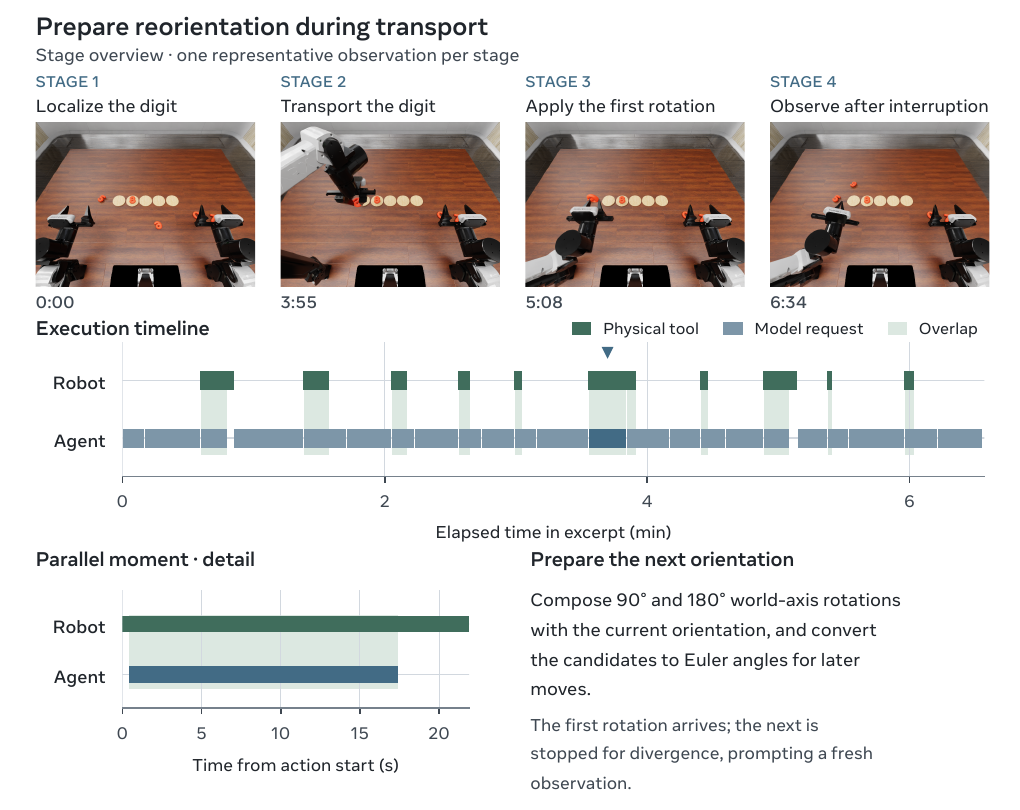}
  \caption{\textbf{Orientation preparation during digit transport.} The upper row summarizes four stages rather than individual tool calls; the lanes below show recorded execution. The enlarged interval shows a model request preparing rotation candidates while transport continues. The later divergence is retained to show how execution feedback affects the prepared plan. Colors and time conventions match Figure~\ref{fig:pipelining-cases}.}
  \label{fig:pipelining-reorientation-case}
\end{figure}

\subsection{Preparing Reorientation During Transport}
\label{app:case-reorientation}

In \texttt{arrange\_largest\_number}, the agent must place digit-shaped objects on pads to form the largest number. The excerpt in Figure~\ref{fig:pipelining-reorientation-case} follows one digit from localization and grasping through transport and reorientation. While the arm moves toward a raised target, the model generates a program that composes candidate $90^\circ$ and $180^\circ$ world-axis rotations with the current orientation and converts them to Euler angles. This prepares later orientation commands without waiting for the transport result.

The transport first exhausts its step budget, so the agent continues toward the same target before attempting the prepared rotation. The first orientation target is reached, but the next rotation stops for divergence. The agent then raises the hand while retaining its measured orientation and captures a new observation. The sequence illustrates why advance preparation and execution monitoring are complementary: a candidate action can be prepared during an earlier movement, while its application still depends on the state returned by the robot.

\endgroup

\section{Sensitivity to the Number of Rollouts}
\label{app:evaluation-budget}

How many rollouts are needed to obtain a stable aggregate Score? We examine this question using a depth-enabled Qwen3.8-Flash-Next evaluation, comparing smaller samples with the full evaluation while keeping the model, harness, and input conditions fixed.

The pool contains 50 rollouts for each of 54 task variants (2,700 episodes): standard and randomized versions of the 12 generalization tasks, plus the other 30 tasks. All results in this appendix come from this depth-enabled configuration; the main simulation results in Section~\ref{sec:simulation-results} use no depth.

\begin{figure}[htbp]
  \centering
  \includegraphics[width=0.82\linewidth]{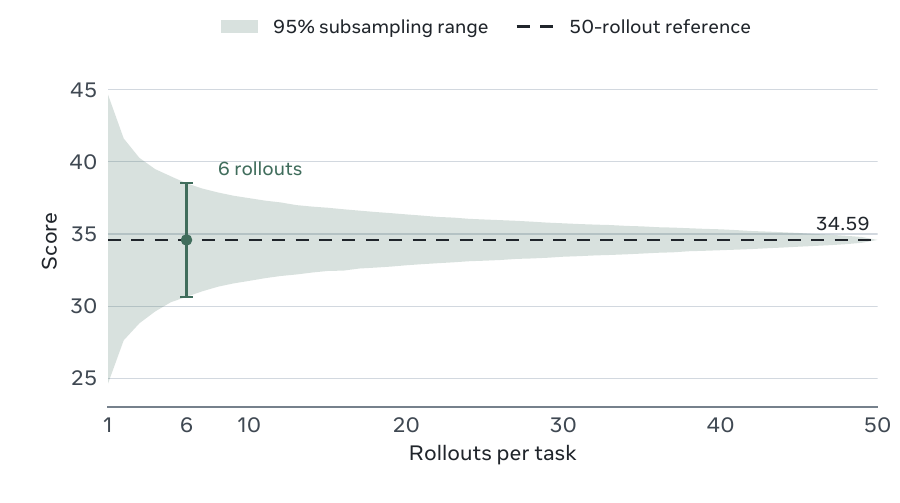}
  \caption{\textbf{Score variability with fewer rollouts.} Shading shows the central 95\% range across 10,000 subsamples of the depth-enabled Qwen evaluation. The dashed line marks the 50-rollout reference; the marker and error bar show the mean and range for six rollouts per task variant.}
  \label{fig:evaluation-budget}
\end{figure}

For each sample size $n=1,\ldots,50$, we independently draw $n$ episodes without replacement from each variant and recompute the Score, averaging within each of the five capability dimensions and then equally across dimensions. We repeat the procedure 10,000 times. Figure~\ref{fig:evaluation-budget} shows the resulting central 95\% range and the full-pool reference of \textbf{34.59}.

With six rollouts per variant, the mean sampled Score is \textbf{34.59}, while the central 95\% range is \textbf{30.65--38.51}. Taking the first six recorded rollouts instead, ordered by pass and layout identifiers, yields \textbf{36.55}, an upward difference of \textbf{1.96 points} from the full-pool reference. Thus, a small evaluation can recover a similar aggregate Score on average while individual subsets still vary by several points. Increasing the number of rollouts reduces this variation. The shaded range describes subsampling from the observed pool and collapses at 50 by construction; it does not directly validate the three-seed, two-layout design used in the main evaluation.

% \section{Additional Real-World Demonstrations}

\end{document}